%% file: main.tex
\documentclass[10pt,twocolumn,letterpaper]{article}

\usepackage[pagenumbers]{cvpr}              %
\usepackage{stfloats}

\input{preamble}

\definecolor{cvprblue}{rgb}{0.21,0.49,0.74}
\usepackage[pagebackref,breaklinks,colorlinks,allcolors=cvprblue]{hyperref}

\def\paperID{11342} %
\def\confName{CVPR}
\def\confYear{2025}

\title{FRAME: Floor-aligned Representation for Avatar Motion from Egocentric Video}

\author{
  Andrea Boscolo Camiletto$^{1,2}$ \quad
  Jian Wang$^{1,2}$ \quad
  Eduardo Alvarado$^{1}$ \quad
  Rishabh Dabral$^{1,2}$ \\
  Thabo Beeler$^{3}$ \quad
  Marc Habermann$^{1,2}$ \quad
  Christian Theobalt$^{1,2}$\\[0.3ex]
  \small
  \textsuperscript{1}Max Planck Institute for Informatics, Saarland Informatics Campus\\
  \small
  \textsuperscript{2}Saarbrücken Research Center for Visual Computing, Interaction and AI~~~~~
  \textsuperscript{3}Google, Switzerland
  \vspace{-3mm}
}

\begin{document}

\input{sec/___title}

\input{sec/00_abstract}

\input{sec/01_intro}
\input{sec/02_related}
\input{sec/03_dataset}
\input{figs/architecture}
\input{sec/04_method}

\input{sec/05_experiments}

\input{sec/10_conclusion}
{
    \small
    \bibliographystyle{ieeenat_fullname}
    \bibliography{references}
}

\end{document}

%% file: preamble.tex
\usepackage{soul}
\usepackage{pifont}
\usepackage{tikz}
\usepackage{tikz-3dplot}
\usepackage{bm}
\usepackage{float}

\usetikzlibrary{positioning, shapes.geometric}
\usetikzlibrary{calc}

\DeclareMathOperator*{\argmin}{arg\,min}

\newcommand{\cmark}{\textcolor{PineGreen}{\ding{51}}}%
\newcommand{\xmark}{\textcolor{BrickRed}{\ding{55}}}%

\newcommand{\Lf}{\mathcal{L}}
\newcommand{\Rf}{\mathcal{R}}
\newcommand{\Mf}{\mathcal{M}}
\newcommand{\Ff}{\mathcal{F}}
\newcommand{\UP}{\uparrow}
\newcommand{\DW}{\downarrow}
\newcommand{\expl}[2]{$\left(\text{\small #1}, #2\right)$}

%% file: sec/___title.tex
\twocolumn[{%
\renewcommand\twocolumn[1][]{#1}%
\maketitle
\centering
\vspace{-5mm}
\includegraphics[width=0.99\textwidth]{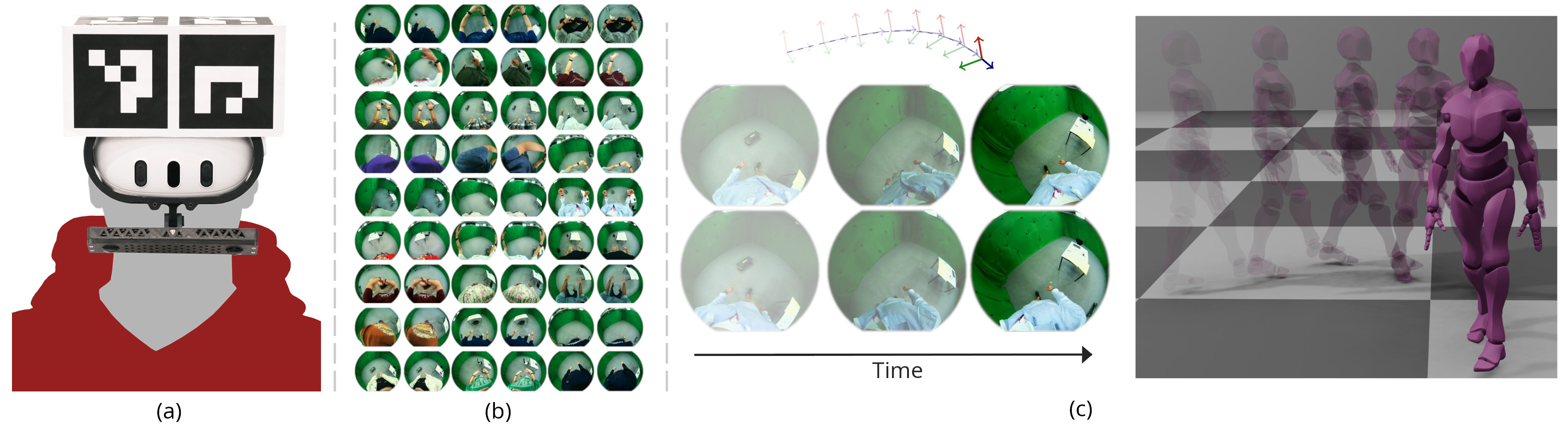}
\captionsetup[figure]{aboveskip=1mm, belowskip=3mm}
\captionof{figure}{We introduce a large scale egocentric dataset (b) collected with a custom-made wearable capture rig (a). With this data we train FRAME, which takes as input a series of egocentric views and the VR pose tracking and predicts the skeletal motion of the user (c).}
\label{fig:teaser}
}]

%% file: sec/00_abstract.tex
\begin{abstract}
\vskip -5mm

Egocentric motion capture with a head-mounted body-facing stereo camera is crucial for VR and AR applications but presents significant challenges such as heavy occlusions and limited annotated real-world data.
Existing methods rely on synthetic pretraining and struggle to generate smooth and accurate predictions in real-world settings, particularly for lower limbs.
Our work addresses these limitations by introducing a lightweight VR-based data collection setup with on-board, real-time 6D pose tracking. Using this setup, we collected the most extensive real-world dataset for ego-facing ego-mounted cameras to date in size and motion variability. 
Effectively integrating this multimodal input -- device pose and camera feeds -- is challenging due to the differing characteristics of each data source.
To address this, we propose FRAME, a simple yet effective architecture that combines device pose and camera feeds for state-of-the-art body pose prediction through geometrically sound multimodal integration and can run at 300 FPS on modern hardware.
Lastly, we showcase a novel training strategy to enhance the model's generalization capabilities.
Our approach exploits the problem's geometric properties, yielding high-quality motion capture free from common artifacts in prior works. 
Qualitative and quantitative evaluations, along with extensive comparisons, demonstrate the effectiveness of our method. Data, code, and CAD designs will be available at \href{https://vcai.mpi-inf.mpg.de/projects/FRAME/}{vcai.mpi-inf.mpg.de/projects/FRAME}.

\end{abstract}

%% file: sec/01_intro.tex
\section{Introduction}
\label{sec:intro}
Egocentric motion capture poses unique challenges, as predicting body pose without an external viewpoint introduces ambiguities and limits contextual information. Nonetheless, the demand for egocentric motion capture spans numerous applications, including virtual reality (VR), augmented reality (AR), remote collaboration, and robotic control.

In VR, body pose estimation traditionally relies on inverse kinematics~\cite{finalik}, using only head and hand poses from the headset and controllers. However, recent trends in industry and research reflect a push towards more accurate body tracking by leveraging priors on human body motion~\cite{agl, avatarposer, questsim, hmd-nemo, egoposer}, using existing forward-facing cameras~\cite{oculus_generative_legs_2023, egoego, egopose, hmd2, egobody3m}, or incorporating additional sensing modalities, such as dedicated body-facing cameras~\cite{apple_vision_pro_2023, egowholebody, unrealego2, egocap}, multiple inertial measurement units (IMU)~\cite{pip, tip, dip, dynaip, dipimu, imuposer, egolocate}, direct time-of-flight sensors~\cite{xrmbt} and pressure-sensing insoles~\cite{soleposer}.

Among wearable alternatives, capturing a user motion via an ego-facing head-mounted camera offers compelling advantages: it provides a top-down view of the user's body and surroundings, and captures detailed cues -- essential for realistic 3D avatars -- including clothing and wrinkles~\cite{chen2024egoavatar}.

For these reasons, many works have tackled this problem, both in monocular~\cite{egowholebody, mo2cap2, xr-egopose, egoglobal, scene-aware, egopw} and stereo~\cite{egocap, egoglass, unrealego, unrealego2, egotap, ego3dpose, simpleego, egoposeformer} settings, although they assume impractical sensor configurations, such as fisheye cameras mounted on protruding bases (see Fig. \ref{fig:rigs}), which are not feasible for actual consumer devices.
In this setting, recent works have introduced large-scale synthetic datasets \cite{egowholebody,mo2cap2,xr-egopose,unrealego,unrealego2}, motion priors \cite{egowholebody}, and additional sensing modalities \cite{totalcapture}. While these developments have significantly improved model capabilities, common limitations persist: poor generalization to in-the-wild data, temporal inconsistencies, and artifacts like body-floor penetration, and foot skating. 

These limitations arise from two primary challenges: significant occlusions in the ego-view and a scarcity of real-world training data. While synthetic datasets have been instrumental, the heavy reliance on them has led to difficulties in generalizing to real-world scenarios as the domain gap is hard to bridge. The limited size of real-world datasets further constrains generalization capabilities of these models.

In this paper, we address these limitations at the very core. We introduce a large-scale dataset that is $\times6$ bigger than what is currently available in the field, eliminating the need for synthetic data pretraining and paving the way for in-the-wild generalization. 
This dataset is captured using a camera positioning that closely reflects a real VR scenario, offering insights into achievable body tracking in a controller-less setup.
An overview of the dataset and the recording rig can be seen in Fig. \ref{fig:teaser}.

Although lightweight SLAM algorithms \cite{orb3} are now widely integrated into everyday devices -- such as VR headsets, robotic vacuum cleaners, and drones -- current egocentric motion capture methods rely on general techniques and often overlook unique setup characteristics such as the known relative pose of the two ego-facing cameras and the device pose provided by these tracking pipelines.

To address this, we design a real-time model that can take advantage of both the camera feeds and the device 6D pose, and instead of implicitly learning their relationship with the user pose, explicitly leverages the specifics of the egocentric setup in a differentiable way. This model achieves state-of-the-art accuracy while running at 300 FPS on modern consumer hardware.
We also introduce a training strategy that effectively combats overfitting and significantly enhances model generalization to unseen data. \\
In summary, this paper presents the following contributions:

\begin{itemize}
    \item A lightweight sensing setup with stereo ego-mounted cameras and head tracking using on-device computations.
    \item An egocentric benchmark dataset with significantly greater scale and motion diversity than existing datasets.
    \item An egocentric motion capture architecture that explicitly leverages the setup geometry, improving MPJPE by 28\% over state of the art and enabling high frame rates.
    \item A training strategy that substantially enhances model generalization to unseen data.
\end{itemize}

%% file: sec/02_related.tex
\section{Related Work}
\label{sec:related}

\par \textbf{Motion Capture using a Single Egocentric Camera.}
One line of work focuses on using a single fisheye camera. xR-EgoPose \cite{xr-egopose}, for instance, utilizes a dual-branch auto-encoder to estimate 3D poses from 2D heatmaps. Selfpose \cite{selfpose} extended this approach by incorporating a joint rotation loss and refining the backbone model. Mo$^2$Cap$^2$ adopted a similar method, training a model to unproject 2D predictions into 3D space.
Recently, Wang \textit{et al.} \cite{scene-aware} and EgoWholeBody \cite{egowholebody} improved the techniques by moving predictions into a 3D volumetric space and a pixel-aligned 3D space, respectively.
Although single-camera setups enable valuable 3D pose estimations, they lack sufficient context for accurate depth estimation, making accurate spatial capture inherently challenging. Therefore, stereo setups are often preferred as they provide enhanced depth information and improve spatial accuracy.

\par \textbf{Motion Capture using an Egocentric Stereo Camera.}
EgoGlass \cite{egoglass} employed a UNet to predict 2D heatmaps, followed by an autoencoder to lift the stereo heatmaps into 3D space. UnrealEgo \cite{unrealego} took a similar approach but utilized a multi-branch autoencoder for 3D pose estimation. UnrealEgo2 \cite{unrealego2} expanded on this by adding segmentation masks, depth prediction, structure-from-motion, and temporal refinement modules. More recent works leverage the kinematic structure of joints, such as EgoTAP \cite{egotap}, which uses a propagation network, and Ego3DPose \cite{ego3dpose}, which compensates for limb size disparities by considering their angle relative to the camera. EgoPoseFormer \cite{egoposeformer} introduced a Pose Refinement Transformer that refines 3D estimates attending visual features in the image space.

Existing methods estimate 3D poses in either the camera coordinate system or relative to the pelvis. Camera-frame estimation benefits from known camera model parameters but lacks broader context, such as gravity-aligned axes. On the other hand, a pelvis-relative estimation cannot leverage the camera's intrinsic prior and needs an estimate of the pelvis position to be aligned with a global frame. Our approach improves on both methods by first predicting in the camera frame and then rototranslating to a floor-aligned reference frame for further refinement. This approach establishes a stable, environment-aligned reference frame that enhances lower-body accuracy and yields more realistic motion capture results.

\par \textbf{Datasets for Egocentric Motion Capture.}
Recent years have seen the release of large-scale datasets in fields such as 3D human pose estimation \cite{4dhumans, h36m} and egocentric action recognition \cite{egoexo4d, ego4d}, enabling significant advancements. However, in egocentric motion capture, the availability of large-scale datasets remains limited.
Collecting large-scale, high-quality datasets for this task is challenging due to the need for specialized devices for markerless motion tracking and complex setups for camera movement tracking.

As a result, most works in this domain rely on synthetic datasets, which provide controlled conditions and readily available ground truth. Mo$^2$Cap$^2$ \cite{mo2cap2} introduced a dataset of 530k images rendered from two cameras, while xR-egopose \cite{xr-egopose} produced 383k images from a single camera. UnrealEgo \cite{unrealego} improved upon this with higher-quality assets and a more advanced rendering pipeline, generating 900k images from two cameras at 25 fps. More recently, UnrealEgo2 \cite{unrealego2} expanded on this with more complex environments and increased the dataset size to 2.5M images.
While synthetic datasets facilitate large-scale data generation with accurate labels, they suffer from domain gaps. The differences in appearance and motion characteristics between synthetic and real-world data often lead to struggles in generalization when models trained on synthetic data are applied to real-world scenarios.

Real-world data collections \cite{unrealego2, egoglass, xr-egopose, mo2cap2, egocap} remain limited, restricting generalization potential and diminishing benchmark reliability due to small dataset sizes. Among the largest, UnrealEgo-RW collected 260k frames matching the camera position of their synthetic dataset, while EgoGlass collected 170k frames. Both total under 2 hours of data.

Another recurring challenge of dataset collection is the tracking of the camera poses.
Some previous works lack camera tracking, limiting evaluations to alignment-based metrics like Procrustes analysis \cite{egopw}. When pose tracking is available, it often requires a cumbersome head-mounted checkerboard \cite{scene-aware, unrealego2, mo2cap2}, which restricts user movement and limits recording duration, leading to unnatural and constrained motion. Moreover, although a checkerboard setup can provide accurate ground-truth device poses, it is unsuitable as input for any method because it does not represent realistic usage scenarios.
Consequently, some approaches rely on computationally intensive techniques, such as SLAM coupled with segmentation pipelines to mask out the human body and estimate camera motion \cite{unrealego2, egoglobal}, resulting in unrealistic head pose tracking incompatible with real-time applications that still lack important details like the ground level.

In contrast, our approach tracks the device pose both from external cameras using a lightweight 3D ArUco board and from onboard sensing directly from a VR device, providing realistic input for our model while also being able to align ground truth labels. This allows us to leverage real-time camera pose information, significantly simplifying the process and improving the reliability of our predictions. 

%% file: sec/03_dataset.tex
\section{SELF Dataset}
To address the lack of real-world data in egocentric motion capture, we introduce a comprehensive dataset focused on body-facing stereo camera setups.
Our dataset captures authentic, diverse, and challenging human movements at unprecedented scale, incorporating environmental interactions absent from prior synthetic or real-world collections.
\subsection{Capture System and Recording Rig}
We captured the dataset in a recording studio equipped with 120 synchronized and calibrated 4K RGB cameras, tracking the skeletal motion with markerless motion capture~\cite{captury}.

\begin{figure}[t]
    \centering
    \begin{tabular}{cc}
        \subfloat[EgoCap \cite{egocap}]{\includegraphics[width=1.5in]{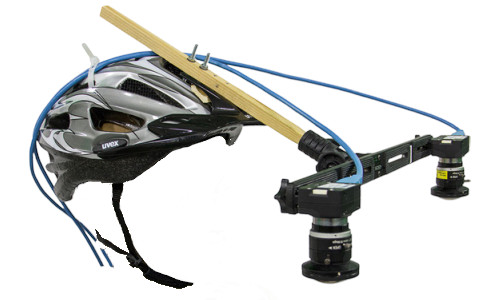}} &
        \subfloat[EgoScene \cite{scene-aware}]{\includegraphics[width=1.5in]{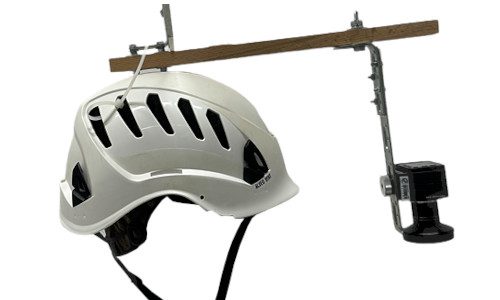}} \\
        \vspace{0.1em} \\
        \subfloat[UnrealEgo2 RW \cite{unrealego2}]{\includegraphics[width=1.5in]{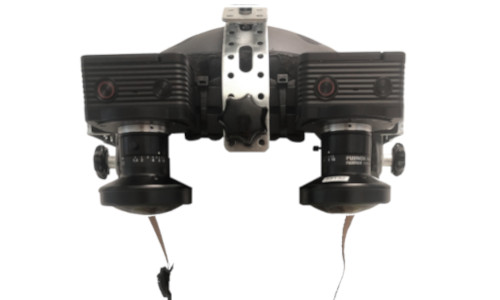}} &
        \subfloat[Ours]{\includegraphics[width=1.5in]{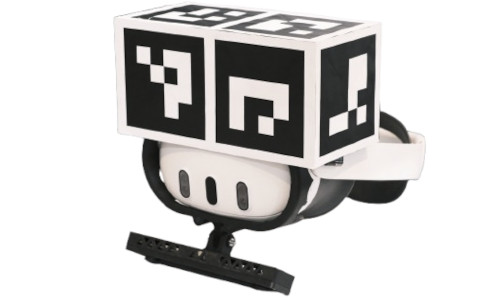}}
    \end{tabular}
    \caption{Comparison of collection devices. With (a), (b), (c) an additional checkerboard must be mounted on top to align ground truth labels, making the helmet significantly top-heavy. Our device can be used for longer period of times and has a camera positioning that mimicks the most a realistic VR scenario.}
   \label{fig:rigs} 
   \vspace{-4mm}
\end{figure}

As existing VR devices lack built-in egocentric cameras or restrict access to their video feeds, we designed a rig centered on the Meta Quest 3 \cite{metaquest3}. Unlike previous setups that relied on helmets outfitted with unrealistically protruding cameras and large checkerboards, our design is lightweight and includes a downward-facing stereo fisheye camera that closely match a realistic VR scenario (see also Fig.~\ref{fig:rigs})

Our rig captures two video streams at 640$\times$480 resolution and 30Hz. The 6D head pose is computed on-device using Quest internal SLAM algorithm. A 3D-printed ArUco board with six markers is mounted on the device solely during dataset collection to allow for ground truth alignment in the VR frame of reference, but is not used during inference.

\subsection{Alignment and Synchronization}\label{sec:align}
To ensure spatial alignment and temporal synchronization across studio cameras, egocentric cameras, and VR onboard pose tracking, we employ a calibration process.

\par
\noindent \textbf{Calibration and Synchronization.}
Each session starts with a one-minute calibration sequence in which participants perform basic motions to establish markerless capture rigging and to calculate transformations between the cameras, markers, and headset using an external checkerboard visible to both the headset and studio cameras. For synchronization, we toggle lights on and off to align the 30Hz fisheye camera clocks with the studio cameras.

\par
\noindent \textbf{Headset Tracking Alignment.} 
For each external camera $i$, we detect any visible ArUco marker $n$ as $\mathbf{d}^i_{n,t} \in \mathbb{R}^{4 \times 2}$ at time $t$. With known marker dimensions and placement, and using the camera extrinsics $\mathbf{P}_i \in \mathbb{R}^{3 \times 4}$ and intrinsics $\mathbf{K}_i \in \mathbb{R}^{3 \times 3}$, we estimate the 6D pose of the ArUco board by solving a Perspective-n-Point problem for each camera, averaging contributions across all cameras.

To refine this estimate, we solve a global optimization problem for each frame, minimizing the sum of reprojection errors across all cameras:
\begin{equation}
\argmin_{\mathbf{N}_t \in \text{SE}(3)} \sum_{i=1}^{C} \sum_{n=1}^{6} \left\| \mathbf{d}^i_{n,t} - \pi\left( \mathbf{K}_i \mathbf{P}_i \mathbf{N}_t \mathbf{X}_n^{\text{hom}} \right) \right\|^2 \forall t,
\end{equation}
where $\mathbf{N}_t$ represents the ArUco board pose at time $t$, $C$ is the number of external cameras, $\mathbf{X}_n^{\text{hom}}$ is the known 3D position of marker $n$ in local homogeneous coordinates, and $\pi(\cdot)$ denotes the perspective projection function.

\par
\noindent \textbf{On-Device Pose Alignment.} 
The VR on-device tracking provides a 6D pose independently of the studio cameras, yet in a different coordinate system and on a separate clock. To align this data with the studio reference, we note that, under ideal conditions, the VR-estimated pose should match the computed 3D ArUco Board pose once we account for the time offset, coordinate system differences, and fixed transformation between the headset and the ArUco board. Hence, we solve the following minimization problem:
\begin{equation}
\argmin_{\mathbf{T}_c,\mathbf{T}_r,t_0} \sum_t^T \left\| \mathbf{N}_t - \mathbf{T}_c \mathbf{V}_{t+t_0} \mathbf{T}_r \right\|^2,
\end{equation}
where $\mathbf{V}_{t+t_0}$ is the VR recorded device pose at internal clock $t$, $t_0
$ represents the clock offset, $\mathbf{T}_c$ is the transformation mapping the VR coordinate system to the studio frame, and $\mathbf{T}_r$ is the fixed transformation between the ArUco board and VR frames. The summation happens over the time dimension. 
We solve this problem with a two-stage optimization: Levenberg-Marquardt \cite{levenberg1944} for transformations and iterative grid search for the clock offset.
The result of this optimization allows us to obtain the device tracked pose in the same frame of reference with our ground truth labels.

\subsection{Dataset Structure}
Our data collection involved 14 participants, each completing two sessions with distinct clothing to enhance visual diversity. Sessions begin with a one-minute calibration, followed by 50 predefined actions (20 seconds each, with 4-second pauses) spanning a variety of sports and daily activities to ensure diverse motions.
Participants can see and interact with their surroundings thanks to the forward-facing cameras on the VR device.

Each session captures stereo egocentric video, 6D device pose, body poses via markerless motion capture, and feeds from a 120-camera studio setup, resulting in over 7 hours and approximately 1.6 million individual images. A separate test set, recorded with two additional participants, offers about an hour of data for unbiased evaluation and model generalization testing.
Notably, our dataset is significantly larger than competing real-world datasets. A detailed comparison is provided in Tab.~\ref{tab:dataset}. 

\input{tables/dataset}

In summary, each time-frame $t$ includes two body-facing fisheye images $\mathbf{I}_t \in \mathbb{R}^{2 \times 3 \times H \times W}$, the onboard device pose $\mathbf{T}_\text{D}(t) \in SE(3)$, and ground truth joint positions $\mathbf{J}_t \in \mathbb{R}^{J \times 3}$.
As is the case with VR, device poses $\mathbf{T}_\text{D}(t)$ are floor-aligned by design -- a critical factor that underpins the predictive methods presented in the following section.

%% file: tables/dataset.tex
\begin{table}[t]
\centering
\resizebox{\columnwidth}{!}{%
\begin{tabular}{lccccccc} 
\toprule

Dataset                         & Cams   & Hrs         &  Frames    & VR Trk      & Act      \\
\hline
EgoCap \cite{egocap}            & 2      & 0.7         &   75k      & \xmark      & 8        \\
Mo2Cap2 \cite{mo2cap2}          & 1      & \textsc{na} &   5k       & \xmark      & 3        \\
xR-EP \cite{xr-egopose}         & 1      & \textsc{na} &   10k      & \xmark      & \textsc{na} \\
EgoGlobal \cite{egoglobal}      & 1      & \textsc{na} &   12k      & \xmark      & 2        \\
SceneEgo \cite{scene-aware}     & 1      & \textsc{na} &   92k      & \xmark      & 5        \\
EgoGlass \cite{egoglass}        & 2      & 1.6         &   173k     & \xmark      & 10       \\
UE-RW \cite{unrealego2}         & 2      & 1.4         &   260k     & \xmark      & 16       \\
\hline
\textbf{Ours}                   & 2      & \bf{7.4}    & \bf{1.6M}  & \cmark      & 14       \\
\bottomrule
\end{tabular}%
}
\caption{Comparison across available egocentric human datasets that provide ground truth annotations. \textit{Cams} stands for Egocentric Cameras, \textit{Hrs} for Hours of video, \textit{VR Trk} for on-device head tracking, \textit{Act} for Actors. Note that our dataset by far outperforms prior datasets in terms of scale while also providing ground truth VR tracking and stereo camera images.}
\label{tab:dataset}
\vspace{-4mm}
\end{table}

%% file: figs/architecture.tex
\begin{figure*}[ht]
    \begin{minipage}[t]{0.41\textwidth}
        \centering
        \resizebox{\textwidth}{!}{\input{figs/architecture-frame}}
        \subcaption{}
        \label{fig:frame}
    \end{minipage}%
    \hspace{0.015\textwidth}%
    \begin{tikzpicture}
        \draw[dashed, gray, line width=0.3pt] (0,0) -- (0,2.6);
    \end{tikzpicture}%
    \hspace{-0.005\textwidth}%
    \begin{minipage}[t]{0.17\textwidth}
        \centering
        \resizebox{\textwidth}{!}{\input{figs/architecture-transform}}
        \subcaption{}
        \label{fig:transform}
    \end{minipage}%
    \hspace{0.01\textwidth}%
    \begin{tikzpicture}
        \draw[dashed, gray, line width=0.3pt] (0,0) -- (0,2.6);
    \end{tikzpicture}%
    \begin{minipage}[t]{0.4\textwidth}
        \centering
        \resizebox{\textwidth}{!}{\input{figs/architecture-time}}
        \subcaption{}
        \label{fig:time}
    \end{minipage}
    \vspace{-8pt}
    \caption{Overview of the proposed architecture. In Fig. \ref{fig:frame}, the backbone accepts two images from the stereo camera as input and outputs two poses, one for each frame. As shown in Fig. \ref{fig:transform} these poses are rototranslated in the frame of reference $\Ff$. In Fig. \ref{fig:time} our STF model accepts the history of two poses from the backbone, aligned in the most recent $\Ff$, and merges the result into a single sequence.}
    \vspace{-4mm}
\end{figure*}
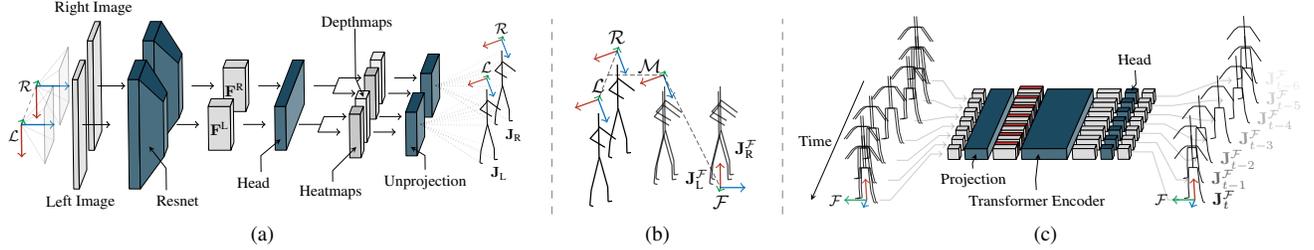

%% file: figs/architecture-frame.tex
\tdplotsetmaincoords{80}{4}%

\newcommand{\drawBox}[8][]{%
    
    \coordinate (#2O) at (#3,#4,#5);
    \coordinate (#2A) at ($(#2O)+(0,0,#8)$);
    \coordinate (#2B) at ($(#2O)+(#6,0,#8)$);
    \coordinate (#2C) at ($(#2O)+(#6,0,0)$);
    \coordinate (#2D) at ($(#2O)+(0,#7,0)$);
    \coordinate (#2E) at ($(#2O)+(0,#7,#8)$);
    \coordinate (#2F) at ($(#2O)+(#6,#7,#8)$);
    \coordinate (#2G) at ($(#2O)+(#6,#7,0)$);

    \coordinate (#2CB) at ($(#2O)!0.5!(#2E)$);
    \coordinate (#2CF) at ($(#2C)!0.5!(#2F)$);
    
    \begin{scope}[line cap=round]

    \draw[fill=#1!80, thin] (#2O) -- (#2A) -- (#2B) -- (#2C) -- cycle;
    \draw[fill=#1!100, thin] (#2A) -- (#2E) -- (#2F) -- (#2B) -- cycle;
    \draw[fill=#1!60, thin] (#2C) -- (#2G) -- (#2F) -- (#2B) -- cycle;

    \end{scope}
}%

\newcommand{\drawTruncatedPyramid}[9][]{%

    \def\tempa{#1}%
    \def\tempb{#2}%
    \def\tempc{#3}%
    \def\tempd{#4}%
    \def\tempe{#5}%
    \def\tempf{#6}%
    \def\tempg{#7}%
    \def\temph{#8}%
    \def\tempi{#9}%

    \drawTruncatedPyramidContinued{\tempb}
}%

\newcommand{\drawTruncatedPyramidContinued}[2]{%

    \def\tempj{#2}%

    \coordinate (#1baseO) at (\tempc,\tempd,\tempe);
    \coordinate (#1baseA) at ($(#1baseO) + (0,0,\tempg)$);
    \coordinate (#1baseB) at ($(#1baseO) + (0,\tempf,\tempg)$);
    \coordinate (#1baseC) at ($(#1baseO) + (0,\tempf,0)$);

    \coordinate (#1topO) at ($(#1baseO) + (\temph, { (\tempf - \tempi ) / 2 }, { (\tempg - \tempj ) / 2 })$);
    \coordinate (#1topA) at ($(#1topO) + (0,0,\tempj)$);
    \coordinate (#1topB) at ($(#1topO) + (0,\tempi,\tempj)$);
    \coordinate (#1topC) at ($(#1topO) + (0,\tempi,0)$);

    \coordinate (#1CB) at ($(#1baseC)!0.5!(#1topC)$);
    \coordinate (#1CF) at ($(#1topO)!0.5!(#1topB)$);

    \fill[\tempa!60] (#1baseC) -- (#1baseO) -- (#1topO) -- (#1topC) -- cycle;
    
    \fill[\tempa!60] (#1topO) -- (#1topA) -- (#1topB) -- (#1topC) -- cycle;
    
    \fill[\tempa!80] (#1baseO) -- (#1baseA) -- (#1topA) -- (#1topO) -- cycle;
    \fill[\tempa!100] (#1baseA) -- (#1baseB) -- (#1topB) -- (#1topA) -- cycle;

    \draw[thin] (#1baseO) -- (#1baseA) -- (#1baseB);
    \draw[thin] (#1topO) -- (#1topA) -- (#1topB) -- (#1topC) -- cycle;
    \draw[thin] (#1baseO) -- (#1topO);
    \draw[thin] (#1baseA) -- (#1topA);
    \draw[thin] (#1baseB) -- (#1topB);
}%

\newcommand{\drawHuman}[7][black]{%
    
    \def\xorigin{#2}
    \def\yorigin{#3}
    \def\zorigin{#4}
    \def\scaleFactor{#5}
    \def\prefix{#6}
    \def\lineWidth{#7}
    
    \def\spineupper{2*\scaleFactor}
    \def\spinelower{-1*\scaleFactor}
    \def\shoulderwidth{0.5*\scaleFactor}
    \def\hipwidth{0.4*\scaleFactor}
    \def\armreach{1*\scaleFactor}
    \def\leglength{3*\scaleFactor}
    
    \coordinate (\prefix-spineTop) at (\xorigin, \yorigin, \zorigin+\spineupper);
    \coordinate (\prefix-spineBottom) at (\xorigin, \yorigin, \zorigin+\spinelower);
    \coordinate (\prefix-shoulderLeft) at (\xorigin, \yorigin+\shoulderwidth, \zorigin+\scaleFactor);
    \coordinate (\prefix-shoulderRight) at (\xorigin, \yorigin-\shoulderwidth, \zorigin+\scaleFactor);
    \coordinate (\prefix-elbowLeft) at (\xorigin+\armreach/2, \yorigin+\armreach, \zorigin+0.5*\scaleFactor);
    \coordinate (\prefix-elbowRight) at (\xorigin-\armreach/2, \yorigin-\armreach, \zorigin+0.5*\scaleFactor);
    \coordinate (\prefix-handLeft) at (\xorigin+\armreach, \yorigin+1.5*\scaleFactor, \zorigin);
    \coordinate (\prefix-handRight) at (\xorigin+\armreach/2, \yorigin-1.5*\scaleFactor, \zorigin);
    \coordinate (\prefix-hipLeft) at (\xorigin, \yorigin+\hipwidth, \zorigin+\spinelower);
    \coordinate (\prefix-hipRight) at (\xorigin, \yorigin-\hipwidth, \zorigin+\spinelower);
    \coordinate (\prefix-kneeLeft) at (\xorigin+\armreach/2, \yorigin+0.6*\scaleFactor, \zorigin+\spinelower-\leglength + \scaleFactor);
    \coordinate (\prefix-kneeRight) at (\xorigin-\armreach/2, \yorigin-0.6*\scaleFactor, \zorigin+\spinelower-\leglength + \scaleFactor);
    \coordinate (\prefix-ankleLeft) at (\xorigin+\armreach/2, \yorigin+0.8*\scaleFactor, \zorigin+\spinelower-\leglength);
    \coordinate (\prefix-ankleRight) at (\xorigin-1.2*\armreach/2, \yorigin-0.6*\scaleFactor, \zorigin+\spinelower-\leglength);
    \coordinate (\prefix-footLeft) at (\xorigin+1.5*\armreach/2, \yorigin+1*\scaleFactor, \zorigin+\spinelower-\leglength);
    \coordinate (\prefix-footRight) at (\xorigin-0.7*\armreach/2, \yorigin-1*\scaleFactor, \zorigin+\spinelower-\leglength);
    
    \draw[line width=\lineWidth, color=#1] (\prefix-spineTop) -- (\prefix-spineBottom);
    
    \draw[line width=\lineWidth, color=#1] (\prefix-shoulderLeft) -- (\prefix-shoulderRight);
    
    \draw[line width=\lineWidth, color=#1] (\prefix-shoulderLeft) -- (\prefix-elbowLeft) -- (\prefix-handLeft);
    \draw[line width=\lineWidth, color=#1] (\prefix-shoulderRight) -- (\prefix-elbowRight) -- (\prefix-handRight);
    
    \draw[line width=\lineWidth, color=#1] (\prefix-hipLeft) -- (\prefix-hipRight);
    
    \draw[line width=\lineWidth, color=#1] (\prefix-hipLeft) -- (\prefix-kneeLeft) -- (\prefix-ankleLeft);
    \draw[line width=\lineWidth, color=#1] (\prefix-hipRight) -- (\prefix-kneeRight) -- (\prefix-ankleRight);
    
    \draw[line width=\lineWidth, color=#1] (\prefix-ankleLeft) -- (\prefix-footLeft);
    \draw[line width=\lineWidth, color=#1] (\prefix-ankleRight) -- (\prefix-footRight);
}%

\newcommand{\drawAxis}[6]{%

  \def\xorigin{#1}
  \def\yorigin{#2}
  \def\zorigin{#3}
  \def\scaleFactor{#4}
  \def\lineWidth{#5}
  \def\prefix{#6}
    
  \coordinate (\prefix-O) at (\xorigin, \yorigin, \zorigin);
  \coordinate (\prefix-X) at (\xorigin+1*\scaleFactor, \yorigin, \zorigin);
  \coordinate (\prefix-Y) at (\xorigin, \yorigin+1*\scaleFactor, \zorigin);
  \coordinate (\prefix-Z) at (\xorigin, \yorigin, \zorigin+1*\scaleFactor);

  \draw[->, color=BrickRed, line width=\lineWidth] (\prefix-O) -- (\prefix-X) node[anchor=north east]{};
  \draw[->, color=ForestGreen, line width=\lineWidth] (\prefix-O) -- (\prefix-Y) node[anchor=north west]{};
  \draw[->, color=NavyBlue, line width=\lineWidth] (\prefix-O) -- (\prefix-Z) node[anchor=south]{};
}%

\newcommand{\drawFrustum}[8]{%

  \coordinate (#1O) at (#2, #3, #4);
  \coordinate (#1A) at ($(#1O) + (#5, -#6, -#7)$);
  \coordinate (#1B) at ($(#1O) + (#5, #6, -#7)$);
  \coordinate (#1C) at ($(#1O) + (#5, #6, #7)$);
  \coordinate (#1D) at ($(#1O) + (#5, -#6, #7)$);

  \draw[opacity=0.1] (#1O) -- (#1A);
  \draw[opacity=0.1] (#1O) -- (#1B);
  \draw[opacity=0.1] (#1O) -- (#1C);
  \draw[opacity=0.1] (#1O) -- (#1D);
  \draw[opacity=0.3, fill=#8!40] (#1A) -- (#1B) -- (#1C) -- (#1D) -- cycle;
}%

\newcommand{\drawCamera}[4]{%

  \coordinate (#1O) at (#2, #3, #4);

  \draw[fill=gray!50] (#1O) -- ++(0.1,0.1,0) -- ++(0,0,0.2) -- ++(-0.1,-0.1,0) -- cycle;
  \draw[fill=gray!50] (#1O) -- ++(0.1,-0.1,0) -- ++(0,0,0.2) -- ++(-0.1,0.1,0) -- cycle;
  \draw[fill=gray!50] (#1O) -- ++(0.1,0.1,0) -- ++(0.1,-0.1,0) -- ++(-0.1,-0.1,0) -- ++(-0.1,0.1,0) -- cycle;
}%

\definecolor{BlockColor}{rgb}{0.1078, 0.288, 0.3784}%
\definecolor{BlockColorLight}{rgb}{0.2078, 0.388, 0.4784}%
\definecolor{BlockColorDark}{rgb}{0.051, 0.105, 0.165}%
\definecolor{TensorColor}{rgb}{0.83, 0.83, 0.83}%
\definecolor{TensorColorLight}{rgb}{0.93, 0.93, 0.93}%
\definecolor{TensorColorDark}{rgb}{0.73, 0.73, 0.73}%

\begin{tikzpicture}[tdplot_main_coords, scale=2.0, line join=bevel, rotate around={0.75:(0,0)}]

  \begin{scope}[shift={(-0.5,0.5,0.5)}]
     \tdplotsetrotatedcoords{0}{90}{0}
     \begin{scope}[tdplot_rotated_coords]
     \drawAxis{0}{0}{0}{0.3}{0.7pt}{axis1-right}
     \end{scope}
  \end{scope}
  \drawFrustum{frustumright}{-0.5}{0.5}{0.5}{0.25}{0.6}{0.25}{TensorColorDark}
  \node[canvas is xz plane at y=0.0,transform shape, anchor=north, scale=0.4] at (-0.6,0.5,0.6) {$\mathcal{R}$};

  \drawBox[TensorColor]{img}{0}{0}{0}{0.05}{1}{1}
  \node[canvas is xz plane at y=0,transform shape, anchor=south, scale=0.4] at (0.025,0.5,1.1) {Right Image};

  \draw[->] ($(imgCF) + (0.0, 0.0, 0.0)$) -- ++(0.25,0,0);

  \drawBox[BlockColor]{enc1}{0.5}{0}{0}{0.1}{1}{1}
  \drawTruncatedPyramid[BlockColor]{enc2}{0.6}{0}{0}{1}{1}{0.2}{0.5}{0.5}

  \draw[->] ($(enc2CF) + (0.1, 0, 0)$) -- ++(0.2,0,0);

  \drawBox[TensorColor]{feats}{1.2}{0.3}{0.3}{0.2}{0.4}{0.4}
  \node[canvas is xz plane at y=0.0,transform shape, anchor=center, scale=0.4] at (1.3,0.3,0.5) {$\mathbf{F}^{\text{R}}$};

  \draw[->] ($(featsCF) + (0.1, 0, 0)$) -- ++(0.2,0,0);
     
  \begin{scope}[opacity=0.1]
    \drawBox[BlockColor]{headright}{1.8}{0.3}{0.3}{0.1}{0.4}{0.4}
  \end{scope}

  \coordinate (P1r) at ($(headrightCF) + (0.1,0,0)$);
  \coordinate (P2r) at ($(P1r) + (0.2,0,0)$);
  \coordinate (P3r) at ($(P2r) + (0,0.4,0)$);
  \coordinate (P4r) at ($(P3r) + (0.2,0,0)$);
  \draw[->] (P1r) -- (P2r) -- (P3r) -- (P4r);

  \coordinate (P6r) at ($(P2r) + (0,-0.4,0)$);
  \coordinate (P7r) at ($(P6r) + (0.2,0,0)$);
  \draw[->] (P2r) -- (P6r) -- (P7r);

  \drawBox[TensorColorLight]{heatmaps}{2.5}{0.7}{0.3}{0.1}{0.4}{0.4}
  \drawBox[TensorColorDark]{depthmaps}{2.5}{-0.1}{0.3}{0.1}{0.4}{0.4}
    
  \draw[->] ($(heatmapsCF) + (0.1, 0, 0)$) -- ++(0.2,0,0);
  \draw[->] ($(depthmapsCF) + (0.1, 0, 0)$) -- ++(0.2,0,0);

  \drawBox[BlockColor]{finalBox}{3.0}{0.0}{0.3}{0.1}{1.0}{0.4}

  \drawHuman{3.7}{0.5}{0.6}{0.11}{humanright}{0.52pt}

  \begin{scope}[shift={($(humanright-spineTop)+(0,0.0,0.1)$)}]
   \tdplotsetrotatedcoords{0}{160}{0}
   \begin{scope}[tdplot_rotated_coords]
     \drawAxis{0}{0}{0}{0.18}{0.5pt}{axis2-right}
   \end{scope}
  \end{scope}
  \node[canvas is xz plane at y=0.0,transform shape, anchor=south, scale=0.4] at ($(humanright-spineTop)+(0,0.0,0.1)$) {$\mathcal{R}$};

  \draw[-, thin, densely dotted, opacity=0.12] (finalBoxCF) -- (humanright-spineTop);
  \draw[-, thin, densely dotted, opacity=0.12] (finalBoxCF) -- (humanright-hipLeft);
  \draw[-, thin, densely dotted, opacity=0.12] (finalBoxCF) -- (humanright-hipRight);
  \draw[-, thin, densely dotted, opacity=0.12] (finalBoxCF) -- (humanright-shoulderLeft);
  \draw[-, thin, densely dotted, opacity=0.12] (finalBoxCF) -- (humanright-shoulderRight);
  \draw[-, thin, densely dotted, opacity=0.12] (finalBoxCF) -- (humanright-elbowLeft);
  \draw[-, thin, densely dotted, opacity=0.12] (finalBoxCF) -- (humanright-elbowRight);
  \draw[-, thin, densely dotted, opacity=0.12] (finalBoxCF) -- (humanright-handLeft);
  \draw[-, thin, densely dotted, opacity=0.12] (finalBoxCF) -- (humanright-handRight);
  \draw[-, thin, densely dotted, opacity=0.12] (finalBoxCF) -- (humanright-kneeLeft);
  \draw[-, thin, densely dotted, opacity=0.12] (finalBoxCF) -- (humanright-kneeRight);
  \draw[-, thin, densely dotted, opacity=0.12] (finalBoxCF) -- (humanright-ankleLeft);
  \draw[-, thin, densely dotted, opacity=0.12] (finalBoxCF) -- (humanright-ankleRight);
  \draw[-, thin, densely dotted, opacity=0.12] (finalBoxCF) -- (humanright-footLeft);
  \draw[-, thin, densely dotted, opacity=0.12] (finalBoxCF) -- (humanright-footRight);

  \def\yoffset{-2}

  \begin{scope}[shift={(-0.5,0.5+\yoffset,0.5)}]
     \tdplotsetrotatedcoords{0}{90}{0}
     \begin{scope}[tdplot_rotated_coords]
       \drawAxis{0}{0}{0}{0.3}{0.7pt}{axis1-left}
     \end{scope}
  \end{scope}
  \drawFrustum{frustumleft}{-0.5}{0.5+\yoffset}{0.5}{0.25}{0.6}{0.25}{TensorColorDark}
  \node[canvas is xz plane at y=\yoffset,transform shape, anchor=north, scale=0.4] at (-0.57,0.5,0.5) {$\mathcal{L}$};

  \draw[dashed, opacity=0.4, dash pattern=on 2pt off 2pt] (frustumrightO) -- (frustumleftO);

  \drawBox[TensorColor]{imgleft}{0}{0+\yoffset}{0}{0.05}{1}{1}
  \node[canvas is xz plane at y=\yoffset,transform shape, anchor=north, scale=0.4] at (0.025,0.5,-0.1) {Left Image};

  \draw[->] ($(imgleftCF) + (0.0, 0.0, 0.0)$) -- ++(0.25,0,0);

  \drawBox[BlockColor]{enc1left}{0.5}{0+\yoffset}{0}{0.1}{1}{1}
  \drawTruncatedPyramid[BlockColor]{enc2left}{0.6}{0+\yoffset}{0}{1}{1}{0.2}{0.5}{0.5}

  \draw[->] ($(enc2leftCF) + (0.1, 0, 0)$) -- ++(0.2,0,0);

  \drawBox[TensorColor]{featsleft}{1.2}{0.3+\yoffset}{0.3}{0.2}{0.4}{0.4}
  \node[canvas is xz plane at y=\yoffset,transform shape, anchor=center, scale=0.4] at (1.3,0.3,0.5) {$\mathbf{F}^{\text{L}}$};

  \draw[->] ($(featsleftCF) + (0.1, 0, 0)$) -- ++(0.2,0,0);
     
  \begin{scope}[opacity=0.1]
    \drawBox[BlockColor]{headleft}{1.8}{0.3+\yoffset}{0.3}{0.1}{0.4}{0.4}
  \end{scope}

  \begin{scope}[opacity=1.0]
  \drawBox[BlockColor]{headmerged}{1.8}{0.3+\yoffset}{0.3}{0.1}{0.4 - \yoffset}{0.4}
  \end{scope}

  \coordinate (P1l) at ($(headleftCF) + (0.1,0,0)$);
  \coordinate (P2l) at ($(P1l) + (0.2,0,0)$);
  \coordinate (P3l) at ($(P2l) + (0,0.4,0)$);
  \coordinate (P4l) at ($(P3l) + (0.2,0,0)$);
  \draw[->] (P1l) -- (P2l) -- (P3l) -- (P4l);

  \coordinate (P6l) at ($(P2l) + (0,-0.4,0)$);
  \coordinate (P7l) at ($(P6l) + (0.2,0,0)$);
  \draw[->] (P2l) -- (P6l) -- (P7l);

  \drawBox[TensorColorLight]{heatmapsleft}{2.5}{0.7+\yoffset}{0.3}{0.1}{0.4}{0.4}
  \drawBox[TensorColorDark]{depthmapsleft}{2.5}{-0.1+\yoffset}{0.3}{0.1}{0.4}{0.4}
    
  \draw[->] ($(heatmapsleftCF) + (0.1, 0, 0)$) -- ++(0.2,0,0);
  \draw[->] ($(depthmapsleftCF) + (0.1, 0, 0)$) -- ++(0.2,0,0);

  \drawBox[BlockColor]{finalBoxleft}{3.0}{0.0+\yoffset}{0.3}{0.1}{1.0}{0.4}

  \drawHuman{3.7}{0.5+\yoffset}{0.6}{0.11}{humanleft}{0.52pt}

  \begin{scope}[shift={($(humanleft-spineTop)+(0,0.0,0.1)$)}]
   \tdplotsetrotatedcoords{0}{160}{0}
   \begin{scope}[tdplot_rotated_coords]
     \drawAxis{0}{0}{0}{0.18}{0.5pt}{axis2-left}
   \end{scope}
  \end{scope}
  \node[canvas is xz plane at y=0.0,transform shape, anchor=south, scale=0.4] at ($(humanleft-spineTop)+(0,0.0,0.1)$) {$\mathcal{L}$};

  \draw[-, thin, densely dotted, opacity=0.12] (finalBoxleftCF) -- (humanleft-spineTop);
  \draw[-, thin, densely dotted, opacity=0.12] (finalBoxleftCF) -- (humanleft-hipLeft);
  \draw[-, thin, densely dotted, opacity=0.12] (finalBoxleftCF) -- (humanleft-hipRight);
  \draw[-, thin, densely dotted, opacity=0.12] (finalBoxleftCF) -- (humanleft-shoulderLeft);
  \draw[-, thin, densely dotted, opacity=0.12] (finalBoxleftCF) -- (humanleft-shoulderRight);
  \draw[-, thin, densely dotted, opacity=0.12] (finalBoxleftCF) -- (humanleft-elbowLeft);
  \draw[-, thin, densely dotted, opacity=0.12] (finalBoxleftCF) -- (humanleft-elbowRight);
  \draw[-, thin, densely dotted, opacity=0.12] (finalBoxleftCF) -- (humanleft-handLeft);
  \draw[-, thin, densely dotted, opacity=0.12] (finalBoxleftCF) -- (humanleft-handRight);
  \draw[-, thin, densely dotted, opacity=0.12] (finalBoxleftCF) -- (humanleft-kneeLeft);
  \draw[-, thin, densely dotted, opacity=0.12] (finalBoxleftCF) -- (humanleft-kneeRight);
  \draw[-, thin, densely dotted, opacity=0.12] (finalBoxleftCF) -- (humanleft-ankleLeft);
  \draw[-, thin, densely dotted, opacity=0.12] (finalBoxleftCF) -- (humanleft-ankleRight);
  \draw[-, thin, densely dotted, opacity=0.12] (finalBoxleftCF) -- (humanleft-footLeft);
  \draw[-, thin, densely dotted, opacity=0.12] (finalBoxleftCF) -- (humanleft-footRight);

  \node[canvas is xz plane at y=0.0,transform shape, anchor=north, scale=0.4] at (0.9,0.5+\yoffset,-0.1) {Resnet};
  \draw[->, >=stealth, solid, line width=0.3pt] (0.8,0.5+\yoffset,-0.1) -- (0.8, 0.5+\yoffset, 0.1) -- (0.65, 0.5+\yoffset, 0.3);

  \node[canvas is xz plane at y=0.0,transform shape, anchor=north, scale=0.4] at (1.6,0.3+\yoffset,0.1) {Head};
  \draw[->, >=stealth, solid, line width=0.3pt] (1.67,0.3+\yoffset,0.1) -- (1.67, 0.3+\yoffset, 0.25) -- (1.85, 0.3+\yoffset, 0.4);

  \node[canvas is xz plane at y=-0.2+\yoffset,transform shape, anchor=west, scale=0.4] at (2.0,0.0,0.0) {Heatmaps};
  \draw[->, >=stealth, solid, line width=0.3pt] (2.35,-0.2+\yoffset,0.1) -- (2.35, -0.2+\yoffset, 0.2) -- (2.56, -0.1+\yoffset, 0.4);

  \node[canvas is xz plane at y=1.2+\yoffset,transform shape, anchor=west, scale=0.4] at (2.12,0.0,1.3) {Depthmaps};
  \draw[->, >=stealth, solid, line width=0.3pt] (2.47,1.2+\yoffset,1.25) -- (2.47, 1.2+\yoffset, 0.8) -- (2.56, 0.9+\yoffset, 0.68);

  \node[canvas is xz plane at y=0.0,transform shape, anchor=north, scale=0.4] at (3.15,0.3+\yoffset,0.12) {Unprojection};
  \draw[->, >=stealth, solid, line width=0.3pt] (3.15,0.3+\yoffset,0.12) -- (3.15, 0.3+\yoffset, 0.25) -- (3.05, -0.1+\yoffset, 0.4);
  \node[canvas is xz plane at y=0.0,transform shape, anchor=east, scale=0.38] at (3.8, 0.2, -0.25) {$\mathbf{J}_{\text{L}}$};
  \node[canvas is xz plane at y=0.0,transform shape, anchor=east, scale=0.38] at (3.8, 0.2-\yoffset, -0.25) {$\mathbf{J}_{\text{R}}$};

\end{tikzpicture}

%% file: figs/architecture-transform.tex
\tdplotsetmaincoords{80}{4}%

\newcommand{\drawHuman}[7][black]{%
    \def\xorigin{#2}
    \def\yorigin{#3}
    \def\zorigin{#4}
    \def\scaleFactor{#5}
    \def\prefix{#6}
    \def\lineWidth{#7}
    \def\spineupper{2*\scaleFactor}
    \def\spinelower{-1*\scaleFactor}
    \def\shoulderwidth{0.5*\scaleFactor}
    \def\hipwidth{0.4*\scaleFactor}
    \def\armreach{1*\scaleFactor}
    \def\leglength{3*\scaleFactor}
    
    \coordinate (\prefix-spineTop) at (\xorigin, \yorigin, \zorigin+\spineupper);%
    \coordinate (\prefix-spineBottom) at (\xorigin, \yorigin, \zorigin+\spinelower);
    \coordinate (\prefix-shoulderLeft) at (\xorigin, \yorigin+\shoulderwidth, \zorigin+\scaleFactor);
    \coordinate (\prefix-shoulderRight) at (\xorigin, \yorigin-\shoulderwidth, \zorigin+\scaleFactor);
    \coordinate (\prefix-elbowLeft) at (\xorigin+\armreach/2, \yorigin+\armreach, \zorigin+0.5*\scaleFactor);
    \coordinate (\prefix-elbowRight) at (\xorigin-\armreach/2, \yorigin-\armreach, \zorigin+0.5*\scaleFactor);
    \coordinate (\prefix-handLeft) at (\xorigin+\armreach, \yorigin+1.5*\scaleFactor, \zorigin);
    \coordinate (\prefix-handRight) at (\xorigin+\armreach/2, \yorigin-1.5*\scaleFactor, \zorigin);
    \coordinate (\prefix-hipLeft) at (\xorigin, \yorigin+\hipwidth, \zorigin+\spinelower);
    \coordinate (\prefix-hipRight) at (\xorigin, \yorigin-\hipwidth, \zorigin+\spinelower);
    \coordinate (\prefix-kneeLeft) at (\xorigin+\armreach/2, \yorigin+0.6*\scaleFactor, \zorigin+\spinelower-\leglength + \scaleFactor);
    \coordinate (\prefix-kneeRight) at (\xorigin-\armreach/2, \yorigin-0.6*\scaleFactor, \zorigin+\spinelower-\leglength + \scaleFactor);
    \coordinate (\prefix-ankleLeft) at (\xorigin+\armreach/2, \yorigin+0.8*\scaleFactor, \zorigin+\spinelower-\leglength);
    \coordinate (\prefix-ankleRight) at (\xorigin-1.2*\armreach/2, \yorigin-0.6*\scaleFactor, \zorigin+\spinelower-\leglength);
    \coordinate (\prefix-footLeft) at (\xorigin+1.5*\armreach/2, \yorigin+1*\scaleFactor, \zorigin+\spinelower-\leglength);
    \coordinate (\prefix-footRight) at (\xorigin-0.7*\armreach/2, \yorigin-1*\scaleFactor, \zorigin+\spinelower-\leglength);
    
    \draw[line width=\lineWidth, color=#1] (\prefix-spineTop) -- (\prefix-spineBottom);
    
    \draw[line width=\lineWidth, color=#1] (\prefix-shoulderLeft) -- (\prefix-shoulderRight);
    
    \draw[line width=\lineWidth, color=#1] (\prefix-shoulderLeft) -- (\prefix-elbowLeft) -- (\prefix-handLeft);
    \draw[line width=\lineWidth, color=#1] (\prefix-shoulderRight) -- (\prefix-elbowRight) -- (\prefix-handRight);
    
    \draw[line width=\lineWidth, color=#1] (\prefix-hipLeft) -- (\prefix-hipRight);
    
    \draw[line width=\lineWidth, color=#1] (\prefix-hipLeft) -- (\prefix-kneeLeft) -- (\prefix-ankleLeft);
    \draw[line width=\lineWidth, color=#1] (\prefix-hipRight) -- (\prefix-kneeRight) -- (\prefix-ankleRight);
    
    \draw[line width=\lineWidth, color=#1] (\prefix-ankleLeft) -- (\prefix-footLeft);
    \draw[line width=\lineWidth, color=#1] (\prefix-ankleRight) -- (\prefix-footRight);
}%

\newcommand{\drawAxis}[6]{%

  \def\xorigin{#1}
  \def\yorigin{#2}
  \def\zorigin{#3}
  \def\scaleFactor{#4}
  \def\lineWidth{#5}
  \def\prefix{#6}
    
  \coordinate (\prefix-O) at (\xorigin, \yorigin, \zorigin);
  \coordinate (\prefix-X) at (\xorigin+1*\scaleFactor, \yorigin, \zorigin);
  \coordinate (\prefix-Y) at (\xorigin, \yorigin+1*\scaleFactor, \zorigin);
  \coordinate (\prefix-Z) at (\xorigin, \yorigin, \zorigin+1*\scaleFactor);

  \draw[->, color=BrickRed, line width=\lineWidth] (\prefix-O) -- (\prefix-X) node[anchor=north east]{};
  \draw[->, color=ForestGreen, line width=\lineWidth] (\prefix-O) -- (\prefix-Y) node[anchor=north west]{};
  \draw[->, color=NavyBlue, line width=\lineWidth] (\prefix-O) -- (\prefix-Z) node[anchor=south]{};
}%

\definecolor{BlockColor}{rgb}{0.1078, 0.288, 0.3784}%
\definecolor{BlockColorLight}{rgb}{0.2078, 0.388, 0.4784}%
\definecolor{BlockColorDark}{rgb}{0.051, 0.105, 0.165}%
\definecolor{TensorColor}{rgb}{0.83, 0.83, 0.83}%
\definecolor{TensorColorLight}{rgb}{0.93, 0.93, 0.93}%
\definecolor{TensorColorDark}{rgb}{0.73, 0.73, 0.73}%

\begin{tikzpicture}[tdplot_main_coords, scale=2.0, line join=bevel, rotate around={0.75:(0,0)}]

  \drawHuman{3.7}{0.5}{0.6}{0.11}{humanright}{0.52pt}

  \begin{scope}[shift={($(humanright-spineTop)+(0,0.0,0.1)$)}]
   \tdplotsetrotatedcoords{0}{160}{0}%
   \begin{scope}[tdplot_rotated_coords]
     \drawAxis{0}{0}{0}{0.18}{0.5pt}{axis2-right}
   \end{scope}%
  \end{scope}%
  \node[canvas is xz plane at y=0.0,transform shape, anchor=south, scale=0.4] at ($(humanright-spineTop)+(0,0.0,0.1)$) {$\mathcal{R}$};

  \def\yoffset{-2}
  \drawHuman{3.7}{0.5+\yoffset}{0.6}{0.11}{humanleft}{0.52pt}%

  \begin{scope}[shift={($(humanleft-spineTop)+(0,0.0,0.1)$)}]
   \tdplotsetrotatedcoords{0}{160}{0}
   \begin{scope}[tdplot_rotated_coords]
     \drawAxis{0}{0}{0}{0.18}{0.5pt}{axis2-left}
   \end{scope}
  \end{scope}
  \node[canvas is xz plane at y=0.0,transform shape, anchor=south, scale=0.4] at ($(humanleft-spineTop)+(0,0.0,0.1)$) {$\mathcal{L}$};

  \draw[dashed, opacity=0.6, dash pattern=on 2pt off 1pt] (axis2-right-O) -- (axis2-left-O);

  \begin{scope}[opacity=0.7]
    \drawHuman{4.1}{0.4+\yoffset/2}{0.6}{0.11}{humanmiddle}{0.52pt}
    \drawHuman{4.1}{0.6+\yoffset/2}{0.6}{0.11}{humanmiddle2}{0.52pt}
  \end{scope}

  \coordinate (midpoint) at ($(humanmiddle-spineTop)!0.5!(humanmiddle2-spineTop)$);
  \begin{scope}[shift={($(midpoint)+(0,0.0,0.1)$)}]
   \tdplotsetrotatedcoords{0}{160}{0}
   \begin{scope}[tdplot_rotated_coords]
     \drawAxis{0}{0}{0}{0.18}{0.5pt}{axis2-middle}
   \end{scope}
  \end{scope}
  \node[canvas is xz plane at y=0.0,transform shape, anchor=south, scale=0.4] at ($(midpoint)+(-0.1,0.0,0.07)$) {$\mathcal{M}$};

  \coordinate (P1) at ($(axis2-right-O)!0.5!(axis2-left-O)$);
  \draw[dashed, opacity=0.6, dash pattern= on 2pt off 1pt] (P1) -- (axis2-middle-O);

  \begin{scope}[opacity=0.7]
    \drawHuman{4.5}{0.4+\yoffset/2}{0.6}{0.11}{humanfloor}{0.52pt}
    \drawHuman{4.5}{0.6+\yoffset/2}{0.6}{0.11}{humanfloor2}{0.52pt}
  \end{scope}

  \begin{scope}[shift={((4.5,0.5+\yoffset/2,0.1))}]
   \tdplotsetrotatedcoords{180}{-90}{0}
   \begin{scope}[tdplot_rotated_coords]
     \drawAxis{0}{0}{0}{0.18}{0.5pt}{axis2-floor}
   \end{scope}
  \end{scope}
  \node[canvas is xz plane at y=0.0,transform shape, anchor=south, scale=0.4] at (4.5,0.5+\yoffset/2,-0.1) {$\mathcal{F}$};

  \draw[dashed, opacity=0.6, dash pattern= on 2pt off 1pt] (axis2-middle-O) -- (axis2-floor-O);
  \node[canvas is xz plane at y=0.0,transform shape, anchor=east, scale=0.35] at (4.6, 0.5-\yoffset, -0.15) {$\mathbf{J}_{\text{R}}^\mathcal{F}$};
  \node[canvas is xz plane at y=0.0,transform shape, anchor=east, scale=0.35] at (4.25, 0.4-\yoffset, -0.35) {$\mathbf{J}_{\text{L}}^\mathcal{F}$};

\end{tikzpicture}

%% file: figs/architecture-time.tex
\tdplotsetmaincoords{70}{10}%
\newcommand{\drawBox}[8][]{%
    
    \coordinate (#2O) at (#3,#4,#5);
    \coordinate (#2A) at ($(#2O)+(0,0,#8)$);
    \coordinate (#2B) at ($(#2O)+(#6,0,#8)$);
    \coordinate (#2C) at ($(#2O)+(#6,0,0)$);
    \coordinate (#2D) at ($(#2O)+(0,#7,0)$);
    \coordinate (#2E) at ($(#2O)+(0,#7,#8)$);
    \coordinate (#2F) at ($(#2O)+(#6,#7,#8)$);
    \coordinate (#2G) at ($(#2O)+(#6,#7,0)$);

    \coordinate (#2CB) at ($(#2O)!0.5!(#2E)$);
    \coordinate (#2CF) at ($(#2C)!0.5!(#2F)$);
    
    \begin{scope}[line cap=round]

    \draw[fill=#1!80, thin] (#2O) -- (#2A) -- (#2B) -- (#2C) -- cycle;
    \draw[fill=#1!100, thin] (#2A) -- (#2E) -- (#2F) -- (#2B) -- cycle;
    \draw[fill=#1!60, thin] (#2C) -- (#2G) -- (#2F) -- (#2B) -- cycle;

    \end{scope}
}%

\newcommand{\drawTruncatedPyramid}[9][]{%

    \def\tempa{#1}%
    \def\tempb{#2}%
    \def\tempc{#3}%
    \def\tempd{#4}%
    \def\tempe{#5}%
    \def\tempf{#6}%
    \def\tempg{#7}%
    \def\temph{#8}%
    \def\tempi{#9}%

    \drawTruncatedPyramidContinued{\tempb}
}%

\newcommand{\drawTruncatedPyramidContinued}[2]{%

    \def\tempj{#2}%

    \coordinate (#1baseO) at (\tempc,\tempd,\tempe);
    \coordinate (#1baseA) at ($(#1baseO) + (0,0,\tempg)$);
    \coordinate (#1baseB) at ($(#1baseO) + (0,\tempf,\tempg)$);
    \coordinate (#1baseC) at ($(#1baseO) + (0,\tempf,0)$);

    \coordinate (#1topO) at ($(#1baseO) + (\temph, { (\tempf - \tempi ) / 2 }, { (\tempg - \tempj ) / 2 })$);
    \coordinate (#1topA) at ($(#1topO) + (0,0,\tempj)$);
    \coordinate (#1topB) at ($(#1topO) + (0,\tempi,\tempj)$);
    \coordinate (#1topC) at ($(#1topO) + (0,\tempi,0)$);

    \coordinate (#1CB) at ($(#1baseC)!0.5!(#1topC)$);
    \coordinate (#1CF) at ($(#1topO)!0.5!(#1topB)$);

    \fill[\tempa!60] (#1baseC) -- (#1baseO) -- (#1topO) -- (#1topC) -- cycle;
    
    \fill[\tempa!60] (#1topO) -- (#1topA) -- (#1topB) -- (#1topC) -- cycle;
    
    \fill[\tempa!80] (#1baseO) -- (#1baseA) -- (#1topA) -- (#1topO) -- cycle;
    \fill[\tempa!100] (#1baseA) -- (#1baseB) -- (#1topB) -- (#1topA) -- cycle;

    \draw[thin] (#1baseO) -- (#1baseA) -- (#1baseB);
    \draw[thin] (#1topO) -- (#1topA) -- (#1topB) -- (#1topC) -- cycle;
    \draw[thin] (#1baseO) -- (#1topO);
    \draw[thin] (#1baseA) -- (#1topA);
    \draw[thin] (#1baseB) -- (#1topB);
}%

\newcommand{\drawHuman}[7][black]{%
    
    \def\xorigin{#2}
    \def\yorigin{#3}
    \def\zorigin{#4}
    \def\scaleFactor{#5}
    \def\prefix{#6}
    \def\lineWidth{#7}
    
    \def\spineupper{2*\scaleFactor}
    \def\spinelower{-1*\scaleFactor}
    \def\shoulderwidth{0.5*\scaleFactor}
    \def\hipwidth{0.4*\scaleFactor}
    \def\armreach{1*\scaleFactor}
    \def\leglength{3*\scaleFactor}
    
    \coordinate (\prefix-spineTop) at (\xorigin, \yorigin, \zorigin+\spineupper);
    \coordinate (\prefix-spineBottom) at (\xorigin, \yorigin, \zorigin+\spinelower);
    \coordinate (\prefix-shoulderLeft) at (\xorigin, \yorigin+\shoulderwidth, \zorigin+\scaleFactor);
    \coordinate (\prefix-shoulderRight) at (\xorigin, \yorigin-\shoulderwidth, \zorigin+\scaleFactor);
    \coordinate (\prefix-elbowLeft) at (\xorigin+\armreach/2, \yorigin+\armreach, \zorigin+0.5*\scaleFactor);
    \coordinate (\prefix-elbowRight) at (\xorigin-\armreach/2, \yorigin-\armreach, \zorigin+0.5*\scaleFactor);
    \coordinate (\prefix-handLeft) at (\xorigin+\armreach, \yorigin+1.5*\scaleFactor, \zorigin);
    \coordinate (\prefix-handRight) at (\xorigin+\armreach/2, \yorigin-1.5*\scaleFactor, \zorigin);
    \coordinate (\prefix-hipLeft) at (\xorigin, \yorigin+\hipwidth, \zorigin+\spinelower);
    \coordinate (\prefix-hipRight) at (\xorigin, \yorigin-\hipwidth, \zorigin+\spinelower);
    \coordinate (\prefix-kneeLeft) at (\xorigin+\armreach/2, \yorigin+0.6*\scaleFactor, \zorigin+\spinelower-\leglength + \scaleFactor);
    \coordinate (\prefix-kneeRight) at (\xorigin-\armreach/2, \yorigin-0.6*\scaleFactor, \zorigin+\spinelower-\leglength + \scaleFactor);
    \coordinate (\prefix-ankleLeft) at (\xorigin+\armreach/2, \yorigin+0.8*\scaleFactor, \zorigin+\spinelower-\leglength);
    \coordinate (\prefix-ankleRight) at (\xorigin-1.2*\armreach/2, \yorigin-0.6*\scaleFactor, \zorigin+\spinelower-\leglength);
    \coordinate (\prefix-footLeft) at (\xorigin+1.5*\armreach/2, \yorigin+1*\scaleFactor, \zorigin+\spinelower-\leglength);
    \coordinate (\prefix-footRight) at (\xorigin-0.7*\armreach/2, \yorigin-1*\scaleFactor, \zorigin+\spinelower-\leglength);
    
    \draw[line width=\lineWidth, color=#1] (\prefix-spineTop) -- (\prefix-spineBottom);
    
    \draw[line width=\lineWidth, color=#1] (\prefix-shoulderLeft) -- (\prefix-shoulderRight);
    
    \draw[line width=\lineWidth, color=#1] (\prefix-shoulderLeft) -- (\prefix-elbowLeft) -- (\prefix-handLeft);
    \draw[line width=\lineWidth, color=#1] (\prefix-shoulderRight) -- (\prefix-elbowRight) -- (\prefix-handRight);
    
    \draw[line width=\lineWidth, color=#1] (\prefix-hipLeft) -- (\prefix-hipRight);
    
    \draw[line width=\lineWidth, color=#1] (\prefix-hipLeft) -- (\prefix-kneeLeft) -- (\prefix-ankleLeft);
    \draw[line width=\lineWidth, color=#1] (\prefix-hipRight) -- (\prefix-kneeRight) -- (\prefix-ankleRight);
    
    \draw[line width=\lineWidth, color=#1] (\prefix-ankleLeft) -- (\prefix-footLeft);
    \draw[line width=\lineWidth, color=#1] (\prefix-ankleRight) -- (\prefix-footRight);
}%

\newcommand{\drawAxis}[6]{%

  \def\xorigin{#1}
  \def\yorigin{#2}
  \def\zorigin{#3}
  \def\scaleFactor{#4}
  \def\lineWidth{#5}
  \def\prefix{#6}
    
  \coordinate (\prefix-O) at (\xorigin, \yorigin, \zorigin);
  \coordinate (\prefix-X) at (\xorigin+1*\scaleFactor, \yorigin, \zorigin);
  \coordinate (\prefix-Y) at (\xorigin, \yorigin+1*\scaleFactor, \zorigin);
  \coordinate (\prefix-Z) at (\xorigin, \yorigin, \zorigin+1*\scaleFactor);

  \draw[->, color=BrickRed, line width=\lineWidth] (\prefix-O) -- (\prefix-X) node[anchor=north east]{};
  \draw[->, color=ForestGreen, line width=\lineWidth] (\prefix-O) -- (\prefix-Y) node[anchor=north west]{};
  \draw[->, color=NavyBlue, line width=\lineWidth] (\prefix-O) -- (\prefix-Z) node[anchor=south]{};
}%

\newcommand{\drawFrustum}[8]{%

  \coordinate (#1O) at (#2, #3, #4);
  \coordinate (#1A) at ($(#1O) + (#5, -#6, -#7)$);
  \coordinate (#1B) at ($(#1O) + (#5, #6, -#7)$);
  \coordinate (#1C) at ($(#1O) + (#5, #6, #7)$);
  \coordinate (#1D) at ($(#1O) + (#5, -#6, #7)$);

  \draw[opacity=0.1] (#1O) -- (#1A);
  \draw[opacity=0.1] (#1O) -- (#1B);
  \draw[opacity=0.1] (#1O) -- (#1C);
  \draw[opacity=0.1] (#1O) -- (#1D);
  \draw[opacity=0.3, fill=#8!40] (#1A) -- (#1B) -- (#1C) -- (#1D) -- cycle;
}%

\newcommand{\drawCamera}[4]{%

  \coordinate (#1O) at (#2, #3, #4);

  \draw[fill=gray!50] (#1O) -- ++(0.1,0.1,0) -- ++(0,0,0.2) -- ++(-0.1,-0.1,0) -- cycle;
  \draw[fill=gray!50] (#1O) -- ++(0.1,-0.1,0) -- ++(0,0,0.2) -- ++(-0.1,0.1,0) -- cycle;
  \draw[fill=gray!50] (#1O) -- ++(0.1,0.1,0) -- ++(0.1,-0.1,0) -- ++(-0.1,-0.1,0) -- ++(-0.1,0.1,0) -- cycle;
}%

\definecolor{BlockColor}{rgb}{0.1078, 0.288, 0.3784}%
\definecolor{BlockColorLight}{rgb}{0.2078, 0.388, 0.4784}%
\definecolor{BlockColorDark}{rgb}{0.051, 0.105, 0.165}%
\definecolor{TensorColor}{rgb}{0.83, 0.83, 0.83}%
\definecolor{TensorColorLight}{rgb}{0.93, 0.93, 0.93}%
\definecolor{TensorColorDark}{rgb}{0.73, 0.73, 0.73}%
\definecolor{EmbColor}{rgb}{0.8, 0.2, 0.2}%

\begin{tikzpicture}[tdplot_main_coords, scale=2.0, line join=bevel, rotate around={3.5:(0,0)}]

  \def\numPoses{6}                    %
  \def\totalWidth{3}                  %
  \def\yAmplitude{0.1}                  %
  \pgfmathsetmacro{\archSize}{1.5}%
  \pgfmathsetmacro{\archStep}{\archSize/\numPoses}%
  \pgfmathsetmacro{\archOffset}{\totalWidth/2 - \archSize/2}%
  \pgfmathsetmacro{\xSpacing}{\totalWidth/\numPoses}%
  \pgfmathsetmacro{\angleStep}{360/\numPoses}%

  \draw[->, >=stealth, solid, line width=0.3pt] (0.3, 0, -0.4) -- (0.3, -\totalWidth, -0.4);
  \node[canvas is xz plane at y=0.0,transform shape, anchor=east, scale=0.4] at (0.3,-\totalWidth/2,-0.4) {Time};

  \begin{scope}[rotate around z=-90, opacity=0.8]
  \foreach \i in {0,1,...,\numPoses} {
    \pgfmathsetmacro{\x}{\xSpacing*\i}                      %
    \pgfmathsetmacro{\y}{\yAmplitude*sin(\i*\angleStep)}    %
    \drawHuman[black]{\x}{0.8+\y}{0}{0.1}{H2\i A}{0.5}         %
    \drawHuman[black]{\x}{0.8+\y-0.02}{0}{0.1}{H2\i B}{0.5}    %

  \ifnum\i=\numPoses
      \begin{scope}[shift={((\x, 0.8+\y-0.01, -0.4))}]
        \tdplotsetrotatedcoords{-90}{90}{180}
        \begin{scope}[tdplot_rotated_coords]
          \drawAxis{0}{0}{0}{0.2}{0.7pt}{axis\i}
        \end{scope}
      \end{scope}
    \fi
  }
  \end{scope}

  \foreach \i in {0,1,...,\numPoses} {
    \pgfmathsetmacro{\y}{\archOffset + \archStep*\i + (\numPoses-\i)/\numPoses* 0.1}
    \drawBox[TensorColor]{E0\i}{1.4}{-\y}{-0.3}{0.1}{0.1}{0.1}
  }

  \foreach \i in {0,1,...,\numPoses} {
    \pgfmathsetmacro{\yH}{\xSpacing*\i}                
    \pgfmathsetmacro{\yT}{\archOffset + \archStep*\i + (\numPoses-\i)/\numPoses* 0.1}
    \begin{scope}[opacity=0.2]
      \draw[->, solid, line width=0.3pt] (0.9,-\yH,-0.4) -- (1.0,-\yH,-0.4) -- (1.25,-\yT,-0.25) -- (1.35,-\yT,-0.25);
    \end{scope}
  }

  \pgfmathsetmacro{\yProj}{-\totalWidth+\archOffset}
  \drawBox[BlockColor]{proj}{1.55}{\yProj}{-0.3}{0.2}{\archSize}{0.1}
  \node[canvas is xz plane at y=0.0,transform shape, anchor=north, scale=0.4] at (1.7,\yProj-0.5,-0.2) {Projection};
  \draw[->, >=stealth, solid, line width=0.3pt] (1.7,\yProj-0.5,-0.2) -- (1.7, \yProj-0.5, -0.15) -- (1.65, \yProj, -0.25);

  \foreach \i in {0,1,...,\numPoses} {
    \pgfmathsetmacro{\y}{\archOffset + \archStep*\i + (\numPoses-\i)/\numPoses* 0.1}
    \drawBox[EmbColor]{E\i}{1.8}{-\y+0.1}{-0.3}{0.2}{0.07}{0.1}
    \drawBox[TensorColor]{E\i}{1.8}{-\y}{-0.3}{0.2}{0.1}{0.1}
  }

  \drawBox[BlockColor]{proj}{2.05}{\yProj}{-0.3}{0.4}{\archSize}{0.1}
  \node[canvas is xz plane at y=0.0,transform shape, anchor=north, scale=0.4] at (2.25,\yProj-0.5,-0.4) {Transformer Encoder};
  \draw[->, >=stealth, solid, line width=0.3pt] (2.25,\yProj-0.5,-0.4) -- (2.25, \yProj-0.5, -0.3) -- (2.2, \yProj, -0.25);

  \foreach \i in {0,1,...,\numPoses} {
    \pgfmathsetmacro{\y}{\archOffset + \archStep*\i + (\numPoses-\i)/\numPoses* 0.1}
    \drawBox[TensorColor]{E\i}{2.5}{-\y}{-0.3}{0.2}{0.1}{0.1}
  }

  \foreach \i in {0,1,...,\numPoses} {
    \pgfmathsetmacro{\y}{\archOffset + \archStep*\i + (\numPoses-\i)/\numPoses* 0.1}
    \drawBox[BlockColor]{E\i}{2.75}{-\y}{-0.3}{0.1}{0.1}{0.1}
  }

  \node[canvas is xz plane at y=0.0,transform shape, anchor=south, scale=0.4] at (2.8,-\archOffset+0.2,-0.1) {Head};
  \draw[->, >=stealth, solid, line width=0.3pt] (2.8,-\archOffset+0.2,-0.1) -- (2.8, -\archOffset+0.2, -0.23) -- (2.8, -\archOffset-0.05, -0.2);

  \foreach \i in {0,1,...,\numPoses} {
    \pgfmathsetmacro{\y}{\archOffset + \archStep*\i + (\numPoses-\i)/\numPoses* 0.1}
    \drawBox[TensorColor]{E\i}{2.9}{-\y}{-0.3}{0.1}{0.1}{0.1}
  }

  \foreach \i in {0,1,...,\numPoses} {
    \pgfmathsetmacro{\yH}{\xSpacing*\i}                
    \pgfmathsetmacro{\yT}{\archOffset + \archStep*\i + (\numPoses-\i)/\numPoses* 0.1 - 0.05}
    \begin{scope}[opacity=0.2]
      \draw[->, solid, line width=0.3pt] (3.0,-\yT,-0.25) -- (3.2,-\yT,-0.25) -- (3.45,-\yH,-0.4) -- (3.55,-\yH,-0.4);
    \end{scope}
  }

  \begin{scope}[rotate around z=-90, opacity=0.9]
  \foreach \i in {0,1,...,\numPoses} {
    \pgfmathtruncatemacro{\idx}{\numPoses-\i}
    \pgfmathsetmacro{\x}{\xSpacing*\i}                      %
    \pgfmathsetmacro{\y}{\yAmplitude*sin(\i*\angleStep)}    %
    \drawHuman[black]{\x}{3.7+\y-0.01}{0}{0.1}{H\i A}{0.5}         %
      \ifnum\idx=0
          \node[canvas is yz plane at x=0.0,transform shape, anchor=west, scale=0.45] at (\x, 3.7+\y-0.01+0.1, -0.4) {$\mathbf{J}^\mathcal{F}_{t}$};
      \else
          \node[canvas is yz plane at x=0.0,transform shape, anchor=west, scale=0.45, opacity=(\i+1)/(\numPoses+4)] at (\x, 3.7+\y-0.01+0.1, -0.4) {$\mathbf{J}^\mathcal{F}_{t-\idx}$};
      \fi

  \ifnum\i=\numPoses
      \begin{scope}[shift={((\x, 3.7+\y-0.01, -0.4))}]
        \tdplotsetrotatedcoords{-90}{90}{180}
        \begin{scope}[tdplot_rotated_coords]
          \drawAxis{0}{0}{0}{0.2}{0.7pt}{axis\i}
        \end{scope}
      \end{scope}
    \fi
  }
  \end{scope}

  \ifnum\i=\numPoses
      \begin{scope}[shift={((\x, 3.7+\y-0.01, -0.4))}]
        \tdplotsetrotatedcoords{-90}{90}{180}
        \begin{scope}[tdplot_rotated_coords]
          \drawAxis{0}{0}{0}{0.2}{0.7pt}{axis\i}
        \end{scope}
      \end{scope}
  \fi
  \node[canvas is xz plane at y=0.0,transform shape, anchor=east, scale=0.4] at (0.62, -\totalWidth, -0.4) {$\mathcal{F}$};
  \node[canvas is xz plane at y=0.0,transform shape, anchor=east, scale=0.4] at (3.5, -\totalWidth, -0.4) {$\mathcal{F}$};
\end{tikzpicture}%

%% file: sec/04_method.tex
\section{Method}
\label{sec:method}
To showcase the effectiveness of our dataset, we propose a method to estimate human pose $\mathbf{J}_t$ at time $t$ from an image stream, $\{\dots, \mathbf{I}_{t-1}, \mathbf{I}_{t}\}$, captured from the head-mounted downward-facing cameras, along with the onboard device poses $\mathbf{T}_{\text{D}}(t)$.

\par
Our approach explicitly leverages this multimodal input, embedding global information in the model during runtime.
First, we utilize the fisheye camera's intrinsic calibrations to make predictions in their local coordinate spaces for the current frame (Sec.~\ref{sec:fisheyepose}). 
Next, we align the current and previous predictions to a common floor and gravity aligned coordinate system using the known relative poses of the cameras and onboard device tracking (Sec.~\ref{sec:coordinates}).
This alignment is crucial, as it enables the model to address challenges such as foot penetration and lower-limb inconsistencies while capturing dynamic motions over time, informed by prior predictions (Sec.~\ref{sec:stereo}).
Our design decisions are motivated by real-time, on-device considerations, enabling the model to run at 300 FPS on an NVIDIA 3090 GPU.

\subsection{Fisheye-based Pose Estimation}\label{sec:fisheyepose}
Building on prior methods on human pose estimation \cite{vnect, xnect, handheatmap, integral}, we train a network to predict a 2.5D representation of keypoints -- comprising 2D image coordinates with their corresponding depth -- and unproject them into 3D space according to our fisheye camera model.
Specifically, given a pair of stereo input images $\mathbf{I}$, we use a pretrained ResNet50 \cite{resnet} to extract visual features $\mathbf{F}^{\text{L}}, \mathbf{F}^{\text{R}} \in\mathbb{R}^{C \times \frac{H}{16} \times \frac{W}{16}}$ for each view, where $H$ and $W$ denote the height and the width of the input image. 
To lower the computational load, we omit the final residual block in the ResNet, obtaining as a side effect an increased spatial resolution that benefits model accuracy.

The features from both cameras are then concatenated and fed to our convolution-based head that predicts, for each camera view and each joint $i$, heatmaps and depthmaps $\mathbf{H}_{i}, \mathbf{D}_{i} \in \mathbb{R}^{\frac{H}{16} \times \frac{W}{16}}$. The heatmaps are then normalized into 2D distributions using a softmax operation, $\mathbf{\hat{H}}_{i} = \operatorname{softmax}(\beta_{i} \cdot \mathbf{H}_{i})$, where $\beta_{i}$ is a learnable temperature parameter. \\
To obtain the $u_{i}, v_{i}$ pixel coordinates, we apply a softargmax operation \cite{softargmax}. We then calculate the depth $d_{i}$ as
\begin{equation}
    d_i = \sum_{j,k}(\mathbf{D}_i \odot \mathbf{\hat{H}}_i)_{jk},
\end{equation}
where $\odot$ denotes the Hadamard product.
More details on the camera model and the unprojection process are provided in the supplementary material.

With $u_i, v_i, d_i$ available for both the views, we unproject each predicted joint to 3D using the fisheye camera model, yielding pose predictions \( \mathbf{J}_{\text{L}} \) and \( \mathbf{J}_{\text{R}} \), where the subscript (L or R) denotes the camera of origin. An overview of this process can be seen in Fig.~\ref{fig:frame}.

\subsection{Coordinate Systems Alignment}\label{sec:coordinates}
Previous methods generally could not assume device pose availability due to hardware constraints, limiting their ability to address the inherent multi-modality of the problem.
In contrast, leveraging the onboard headset pose, we employ a principled approach: rather than learning the relationship between skeletal motion and the device pose, we use it explicitly to move the problem to a shared, floor-aligned frame of reference.
Given the headset pose $\mathbf{T}_\text{D}$ at any point in time and the fixed relative transformations $\mathbf{M}_{\text{L}}, \mathbf{M}_{\text{R}} \in SE(3)$ between the headset and each camera, we can compute the global pose of each camera $\mathbf{T}_{\mathcal{L}}, \mathbf{T}_{\mathcal{R}}$ as follows:
\[
\mathbf{T}_{\mathcal{L}} = \mathbf{T}_\text{D} \cdot \mathbf{M}_{\text{L}}, \quad \mathbf{T}_{\mathcal{R}} = \mathbf{T}_\text{D} \cdot \mathbf{M}_{\text{R}},
\]
Although a global frame provides useful information, such as gravity direction and floor positioning, it remains suboptimal as both its origin and horizontal axes are arbitrary.
To account for this, we define a new reference frame $\Ff$ that is aligned with the floor, as described in Fig.~\ref{fig:frames}.

\input{figs/frame_of_refs}

This allows us to compute the relative transformations, ${}^{\mathcal{F}}\mathbf{T}_\mathcal{L}$ and ${}^\mathcal{F}\mathbf{T}_\mathcal{R}$, of the two cameras to the floor's frame of reference and use them to align each camera’s predicted 3D joints into the common, floor-aligned frame $\mathcal{F}$
\[
\mathbf{J}_L^{\mathcal{F}} = {}^\mathcal{F}\mathbf{T}_\mathcal{L} \cdot \mathbf{J}_L, \quad \mathbf{J}_R^{\mathcal{F}} = {}^\mathcal{F}\mathbf{T}_{\mathcal{R}} \cdot \mathbf{J}_R,
\]

Note that any device slipping is inherently accounted for, as onboard SLAM continuously tracks the actual headset pose, naturally capturing variations in its position relative to the head.

\subsection{Stereo Temporal Fusion}\label{sec:stereo}
\par
To compute the final estimate, $\mathbf{J}_t$, we align the pose pairs from the current and the past steps to the most recent $\mathcal{F}$ as $ \{\dots \,, \mathbf{J}^\Ff_{\text{L},t-1}, \mathbf{J}^\Ff_{\text{R},t-1}, \mathbf{J}^\Ff_{\text{L},t}, \mathbf{J}^\Ff_{\text{R},t}\}$ , yielding a complete motion sequence in a shared coordinate system.
This design provides several advantages.
First, by establishing a stable reference aligned with the ground, we can correct artifacts such as floating feet or ground penetration. 
Additionally, by knowing the direction of gravity, one can compensate for unstable poses or preserve them when consistent with the motion history from previous steps. 

Having the estimated per-view, per-frame body poses in the floor's frame of reference, we finally perform a multi-view temporal fusion of the coarse pose estimates.
Specifically, we input this sequence of pose pairs into our Stereo Temporal Fusion (STF) module, a transformer encoder tailored for this task. %
STF is structured with 8 layers, each having a feedforward dimension of 512 and 32 attention heads, and processes the previous 20 predictions sampled at 15Hz. A visualization can be seen in Fig.~\ref{fig:time}.

The model outputs refined motion by integrating stereo predictions, floor alignment, and temporal coherence. 
It is worth noting that while it predicts the full motion sequence during training, our method operates in real-time at inference, outputting only the most recent frame.

\subsection{Cross-Training Caching}\label{sec:trainstrat}
Image-based pose estimation models typically improve generalization as training progresses, although performance on unseen data naturally lags behind improvements on the training set. This discrepancy negatively impacts downstream modules — such as our Stereo Temporal Fusion (STF) — which, when trained on these overly accurate backbone predictions, struggle to generalize to inputs with more realistic error distributions.
\par
To mitigate this, we introduce a Cross-Training Caching approach inspired by k-fold cross-validation, specifically designed to replicate realistic errors during training.
We divide our training set into $k$ subsets and iteratively re-train our backbone on $k-1$ subsets, caching predictions on the held-out subset. After repeating this process $k$ times, we obtain backbone predictions for the entire training set. These cached predictions, reflecting realistic errors on unseen data, are used exclusively to train the STF module, ensuring it learns robustly from inputs that closely mimic inference-time conditions.
A visualization of this mechanism is shown in Fig.~\ref{fig:training}. 
The final model employs a backbone trained on the entire training set.
\input{figs/training}

%% file: figs/frame_of_refs.tex
\begin{figure}[t]
\centering

\tdplotsetmaincoords{62}{30}%

\begin{tikzpicture}[tdplot_main_coords,  line join=bevel]
    \def\width{5}
    \def\depth{1.0}
    \def\height{0.3}
    
    \def\w{\width/2}
    \def\d{\depth/2}
    \def\h{\height/2}

    \def\offset{0.5}

    \def\l{1.1}
    \def\axisthickness{0.8pt}
    
    \colorlet{xcolor}{BrickRed}
    \colorlet{ycolor}{ForestGreen}
    \colorlet{zcolor}{MidnightBlue}
    \begin{scope}[rotate around y=180, rotate around x=25, rotate around z=0]
      \coordinate (O) at (-\w,-\d,-\h);
      \coordinate (A) at (\w,-\d,-\h);
      \coordinate (B) at (\w,\d,-\h);
      \coordinate (C) at (-\w,\d,-\h);
      \coordinate (D) at (-\w,-\d,\h);
      \coordinate (E) at (\w,-\d,\h);
      \coordinate (F) at (\w,\d,\h);
      \coordinate (G) at (-\w,\d,\h);

      \coordinate (H) at (-\w + 2*\d, -\d, -\h);
      \coordinate (I) at (-\w + 2*\d, \d, -\h);
      \coordinate (J) at (-\w + 2*\d, \d, \h);
      \coordinate (K) at (-\w + 2*\d, -\d, \h);

      \coordinate (L) at (\w - 2*\d, -\d, -\h);
      \coordinate (M) at (\w - 2*\d, \d, -\h);
      \coordinate (N) at (\w - 2*\d, \d, \h);
      \coordinate (P) at (\w - 2*\d, -\d, \h);

      \draw (O) -- (H) -- (I) -- (C) -- cycle;
      \draw (A) -- (L) -- (M) -- (B) -- cycle;

      \draw (O) -- (D) -- (G) -- (C);
      \draw (D) -- (K) -- (H);  

      \draw (M) -- (N) -- (P) -- (L);
      \draw (P) -- (E) -- (A);
  
      \coordinate (Q) at (-\w + 2*\d, 0, \h);
      \coordinate (R) at (\w - 2*\d, 0, \h);
      \draw[dashed, gray, opacity=0.3] (Q) -- (R);

      \coordinate (O1) at (-\w + \offset, 0, \h);
      \coordinate (O2) at (\w - \offset, 0, \h);
      \coordinate (O3) at (0, 0, \h);

      \draw[->, xcolor, line width=\axisthickness] (O1) -- ++(\l,0,0) node[anchor=north east, xcolor]{};
      \draw[->, ycolor, line width=\axisthickness] (O1) -- ++(0,\l,0) node[anchor=north west, ycolor]{};
      \draw[->, zcolor, line width=\axisthickness] (O1) -- ++(0,0,\l) node[anchor=south, zcolor]{};
      \node at ($(O1) + (-0.4, 0.0, 0.3)$) {$\mathcal{L}$};

      \draw[->, xcolor, line width=\axisthickness] (O2) -- ++(\l,0,0) node[anchor=north east, xcolor]{};
      \draw[->, ycolor, line width=\axisthickness] (O2) -- ++(0,\l,0) node[anchor=north west, ycolor]{};
      \draw[->, zcolor, line width=\axisthickness] (O2) -- ++(0,0,\l) node[anchor=south, zcolor]{};
      \node at ($(O2) + (0.4, 0.0, 0.3)$) {$\mathcal{R}$};

      \draw[->, xcolor, line width=\axisthickness] (O3) -- ++(\l,0,0) node[anchor=north east, xcolor]{};
      \draw[->, ycolor, line width=\axisthickness] (O3) -- ++(0,\l,0) node[anchor=north west, ycolor]{};
      \draw[->, zcolor, line width=\axisthickness] (O3) -- ++(0,0,\l) node[anchor=south, zcolor]{};
      \node at ($(O3) + (0.35, 0.0, 0.35)$) {$\mathcal{M}$};

      \def\fl{0.8}
      \def\fh{0.3}
      \def\fw{0.6}

      \coordinate (C2) at ($(O1) + (-\fw,-\fh,\fl)$);
      \coordinate (C3) at ($(O1) + (\fw,-\fh,\fl)$);
      \coordinate (C4) at ($(O1) + (\fw,\fh,\fl)$);
      \coordinate (C5) at ($(O1) + (-\fw,\fh,\fl)$);

      \draw[gray, opacity=0.3] (C2) -- (C3) -- (C4) -- (C5) -- cycle;
      \draw[gray, opacity=0.3] (O1) -- (C2);
      \draw[gray, opacity=0.3] (O1) -- (C3);
      \draw[gray, opacity=0.3] (O1) -- (C4);
      \draw[gray, opacity=0.3] (O1) -- (C5);

      \begin{scope}
          \clip (C2) -- (C3) -- (C4) -- (C5) -- cycle;
          \fill[gray, opacity=0.1] (O1) -- (C2) -- (C3) -- (O1);
          \fill[gray, opacity=0.1] (O1) -- (C3) -- (C4) -- (O1);
          \fill[gray, opacity=0.1] (O1) -- (C4) -- (C5) -- (O1);
          \fill[gray, opacity=0.1] (O1) -- (C5) -- (C2) -- (O1);
      \end{scope}
      
      \coordinate (C2) at ($(O2) + (-\fw,-\fh,\fl)$);
      \coordinate (C3) at ($(O2) + (\fw,-\fh,\fl)$);
      \coordinate (C4) at ($(O2) + (\fw,\fh,\fl)$);
      \coordinate (C5) at ($(O2) + (-\fw,\fh,\fl)$);

      \draw[gray, opacity=0.3] (C2) -- (C3) -- (C4) -- (C5) -- cycle;
      \draw[gray, opacity=0.3] (O2) -- (C2);
      \draw[gray, opacity=0.3] (O2) -- (C3);
      \draw[gray, opacity=0.3] (O2) -- (C4);
      \draw[gray, opacity=0.3] (O2) -- (C5);

      \begin{scope}
          \clip (C2) -- (C3) -- (C4) -- (C5) -- cycle;
          \fill[gray, opacity=0.1] (O2) -- (C2) -- (C3) -- (O2);
          \fill[gray, opacity=0.1] (O2) -- (C3) -- (C4) -- (O2);
          \fill[gray, opacity=0.1] (O2) -- (C4) -- (C5) -- (O2);
          \fill[gray, opacity=0.1] (O2) -- (C5) -- (C2) -- (O2);
      \end{scope}

    \end{scope}

    \def\planeofs{2.5}
    \coordinate (O4) at ($(O3) - (0,0,\planeofs)$);
    \draw[dashed] (O3) -- (O4);

    \def\circleradius{1.4}
    \begin{scope}
        \clip (O4) circle (\circleradius);
        \fill[gray!20,opacity=0.3] (O4) circle (\circleradius);
        \foreach \i in {-2,-1.85,...,2} {
            \draw[gray,opacity=0.1] ($(O4)+(-\circleradius,\i,0)$) -- ($(O4)+(\circleradius,\i,0)$);
        }
    \end{scope}
    \draw (O4) circle (\circleradius);

    \draw[->, xcolor, line width=\axisthickness] (O4) -- ++(\l,0,0) node[anchor=north east, xcolor]{};
    \draw[->, ycolor, line width=\axisthickness] (O4) -- ++(0,0,\l) node[anchor=north west, ycolor]{};
    \draw[->, zcolor, line width=\axisthickness] (O4) -- ++(0,-\l,0) node[anchor=south, zcolor]{};
    \node at ($(O4) + (0.0, 0.35, 0.0)$) {$\mathcal{F}$};

\end{tikzpicture}

\caption[]{$\mathcal{L}$ and $\mathcal{R}$ are the left/right camera frames, $\mathcal{M}$ is the middle frame computed as the average of the two camera frames. The $x, y, z$ axes are color-coded to red, green, and blue, respectively.
$\mathcal{F}$ is obtained by moving the origin of $\mathcal{M}$ at the ground level, aligning $y$ axis to be vertical, and using the projection of the $x$ axis of $\Mf$ on the horizontal plane to determine the direction of the horizontal axes of $\Ff$}

\label{fig:frames}
\vspace{-5mm}
\end{figure}
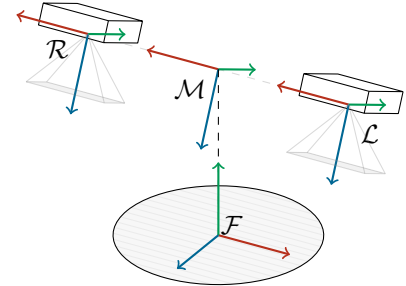

%% file: figs/training.tex
\begin{figure}[t]
\centering{
\resizebox{0.7\columnwidth}{!}{
\begin{tikzpicture}[
    box/.style={draw, rounded corners=1pt, line width=0.4pt},
]

\definecolor{primarycolor}{HTML}{e9c46a}
\definecolor{secondarycolor}{HTML}{925E78}

\pgfmathsetmacro{\h}{0.48}     %
\pgfmathsetmacro{\sp}{0.1}    %
\pgfmathsetmacro{\tw}{3.0}    %

\pgfmathsetmacro{\sb}{\tw/3 - \sp/2}   %
\pgfmathsetmacro{\mb}{\tw * 2/3 - \sp/2}    %
\pgfmathsetmacro{\lb}{3.0}      %
\pgfmathsetmacro{\vs}{\h + \sp}      %
\pgfmathsetmacro{\c}{\tw/2}     %

\node[box, minimum width=\lb cm, minimum height=\h cm, anchor=north west, fill=primarycolor] at (0,0) {\footnotesize Training Data};

\draw[dashed, opacity=0.5] (0,-\vs) -- (\tw,-\vs);

\node[anchor=east] at (-0.4, -\vs-\sp-\h/2) {\scriptsize Run 1};
\node[box, minimum width=\mb cm, minimum height=\h cm, anchor=north west, fill=primarycolor] at (0,-\vs-\sp) {};
\node[box, minimum width=\sb cm, minimum height=\h cm, anchor=north east, fill=secondarycolor] at (\tw,-\vs-\sp) {\footnotesize $H_1$};

\node[anchor=east] at (-0.4, -2*\vs-\sp-\h/2) {\scriptsize Run 2};
\node[box, minimum width=\sb cm, minimum height=\h cm, anchor=north west, fill=secondarycolor] at (0,-2*\vs-\sp) {\footnotesize $H_2$};
\node[box, minimum width=\mb cm, minimum height=\h cm, anchor=north east, fill=primarycolor] at (\tw,-2*\vs-\sp) {};

\node[anchor=east] at (-0.4, -3*\vs-\sp-\h/2) {\scriptsize Run 3};
\node[box, minimum width=\sb cm, minimum height=\h cm, anchor=north west, fill=primarycolor] at (0,-3*\vs-\sp) {};
\node[box, minimum width=\sb cm, minimum height=\h cm, anchor=north, fill=secondarycolor] at (\c,-3*\vs-\sp) {\footnotesize $H_3$};
\node[box, minimum width=\sb cm, minimum height=\h cm, anchor=north east, fill=primarycolor] at (\tw,-3*\vs-\sp) {};

\draw[dashed, opacity=0.5] (0,-4*\vs-\sp) -- (\tw,-4*\vs-\sp);

\node[box, minimum width=\sb cm, minimum height=\h cm, anchor=north west, fill=secondarycolor] at (0,-4*\vs-2*\sp) {\footnotesize $H_2$};
\node[box, minimum width=\sb cm, minimum height=\h cm, anchor=north, fill=secondarycolor] at (\c,-4*\vs-2*\sp) {\footnotesize $H_3$};
\node[box, minimum width=\sb cm, minimum height=\h cm, anchor=north east, fill=secondarycolor] at (\tw,-4*\vs-2*\sp) {\footnotesize $H_1$};

\draw[->, line width=0.5pt] (\tw,-\vs-\sp-\h/2) -- ++(0.3,0) -- (\tw+0.3, -4*\vs-2*\sp-\h/2) -- ++(-0.3,0);
\draw[->, line width=0.5pt] (0,-2*\vs-\sp-\h/2) -- ++(-0.2,0) -- (-0.2, -4*\vs-2*\sp-\h/2) -- ++(0.2,0);
\draw[->, line width=0.5pt] (\tw/2,-3*\vs-\sp-\h) -- (\tw/2,-4*\vs-2*\sp);

\end{tikzpicture}
}}
\caption{Visualization of the k-fold Cross Training Caching Strategy with $k=3$. $H_i$ denotes the hold-out data for the $i$-th run.}
\label{fig:training}
\vspace{-4mm}
\end{figure}
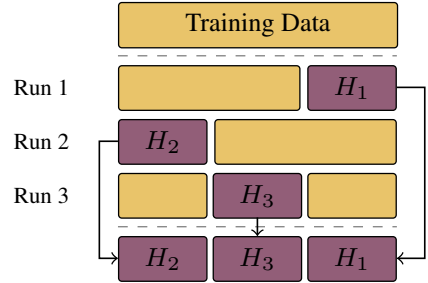

%% file: sec/05_experiments.tex
\section{Experiments} \label{sec:experiments}
\par
\noindent \textbf{Implementation Details.}
All input images are resized to $256\times256$, and we utilize the AdamW optimizer \cite{adamw} with batch size 16 for training.
We first train our backbone fisheye-based pose estimation module with a learning rate of $3\cdot10^{-5}$ and a weight decay $10^{-5}$ for 6 epochs. 
We apply an L2 loss on both predictions of the model in camera coordinates.
We then follow the strategy explained in Sec.~\ref{sec:trainstrat} to train the STF module, with a learning rate of $3\cdot10^{-4}$ and weight decay $10^{-4}$ for 2 epochs with the same loss, but now in global coordinates.
\par
\noindent \textbf{Dataset.}
From the 14 available subjects in the FRAME dataset, we exclude two (one male and one female) from the training set to assess the generalization capability of each model. 
Since the cameras are already aligned with the head, we follow best practices in the field \cite{scene-aware, egowholebody} and do not align the body predictions with the ground truth when evaluating. 
We adopt a 15-keypoint skeleton consistent with the EgoScene dataset \cite{scene-aware} and adapt the baselines to this skeleton where necessary. 
\par
\noindent \textbf{Metrics.}
As in prior works, we report results in terms of mean per joint positional error (MPJPE) in millimeters. 
In addition, we report other complementary metrics, i.e., Procrustes-aligned MPJPE (PA-MPJPE), 3D percentage of correct keypoints (3D-PCK) within a 10cm threshold, Jitter, Non-Penetration Percentage (NPP), Mean Penetration Error (MPE), and Foot Sliding (FS).
A detailed explanation on how these metrics are computed is present in the supplementary materials.
\subsection{Comparison}
To ensure a rigorous comparison with state-of-the-art methods, we retrain three existing methods, i.e. UnrealEgo~\cite{unrealego}, EgoGlass\cite{egoglass}, and EgoPoseFormer\cite{egoposeformer}, on our dataset according to the original settings described in the papers. 
If a method was initially trained on predicting the root-relative pose, we adjust it to output the pose in the camera frame, as it would require knowing ground truth information (the root position) in order to evaluate its MPJPE.
The results in Tab.~\ref{tab:methods} highlight the effectiveness of our approach across multiple key metrics.
Our model design achieves the lowest MPJPE among all the tested methods, while maintaining a lightweight architecture and running at 300 FPS on an NVIDIA RTX 3090, which is significatly faster than prior works.
This performance is particularly noteworthy in metrics related to lower-body alignment, where the integration of device pose tracking enables precise floor alignment and prevents common artifacts such as floor penetration, leading to a perfect Non-Penetration Percentage (NPP) score.
Further, our method showcases improved stability in sequential predictions, as evidenced by reductions in, both, jitter and foot sliding. 
These enhancements are largely attributed to the Stereo Temporal Fusion (STF) module, which incorporates prior predictions to yield temporally smoother motions.
Comparative analysis with baseline methods reveals the substantial benefits of our combined stereo-temporal fusion and floor alignment strategy and overall highlights the advantages of our multi-modal integration, achieving state-of-the-art performance across all evaluated metrics.
\begin{figure*}[t]
    \centering
    \includegraphics[width=0.95\textwidth]{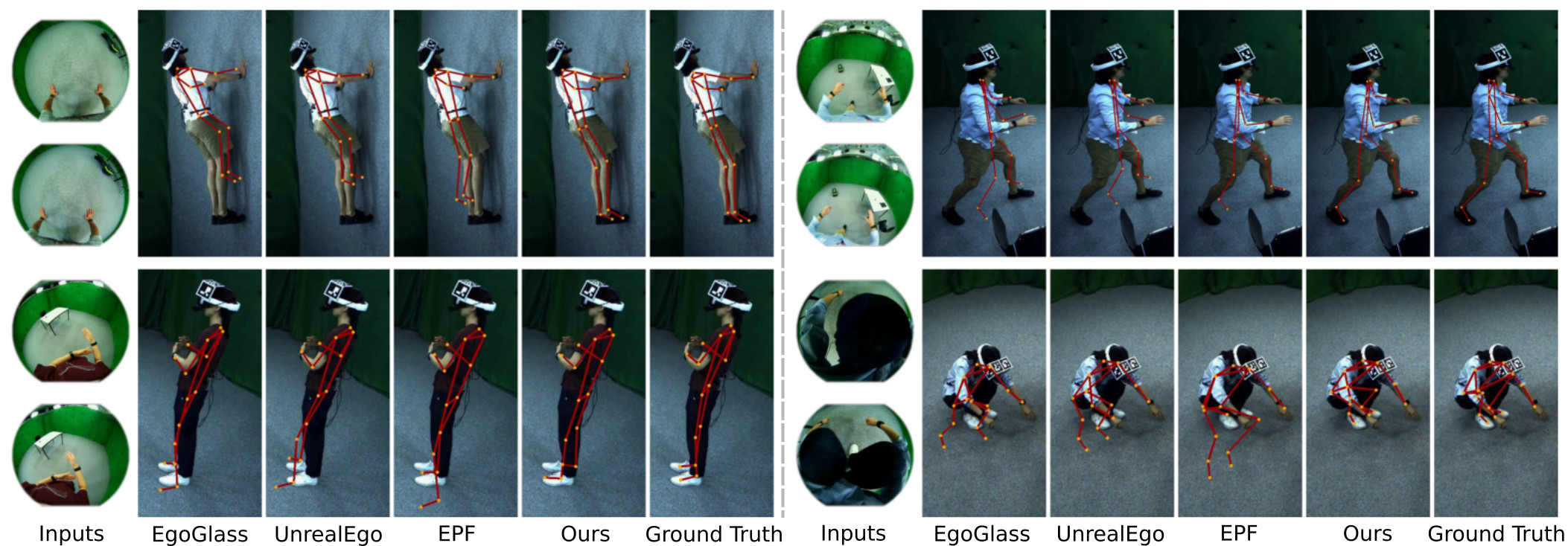}
    \caption{
    Qualitative comparison on challenging inputs. 
    The predicted 3D poses are overlayed onto external reference views not used for tracking. 
    Our qualitative results confirm that our method predicts more accurate body poses, and significantly better handles contacts with the floor and lower limbs compared to prior state-of-the-art approaches~\cite{unrealego, egoglass, egoposeformer}.
    }
    \label{fig:your_label}
    \vspace{-2mm}
\end{figure*}
\input{tables/methods}
\subsection{Ablation Study}
We systematically evaluate the effect of different configurations on the capture performance.
Tab.~\ref{tab:ablation} summarizes our ablation study. 
Each row reflects a distinct experimental setting, incrementally incorporating model components to isolate their contribution. 
To better disentangle the impact of our frame alignment and the use of predictions at previous steps, we report some results where only an MLP head is used for merging the stereo views, without any information on the prediction history. 
\par
\noindent \textbf{Baseline (w/o stereo).} 
When only the left camera view is used without stereo information, the model lacks depth cues provided by binocular views, resulting in the highest MPJPE (87.94 mm).
\par
\noindent \textbf{Stereo Fusion by Averaging (w/ avg).} 
Averaging predictions from the two stereo views, now aligned in the middle frame ($\Mf$), reduces MPJPE to 84.31 mm, due to the partial balancing out of discrepancies in each view prediction.
\par
\noindent \textbf{Learned Stereo Fusion (w/ MLP).} 
Introducing an MLP to merge left and right views without any explicit alignment further decreases MPJPE to 82.19mm. 
This indicates that a learnable fusion can adjust to the different view biases.
Explicit alignment before the MLP provides an additional improvement to 81.69mm.
\par
\noindent \textbf{Shared Head (w/ SH).}
Processing image features jointly rather than independently reduces MPJPE to 74.13mm, suggesting that stereo information can significantly help the model to resolve ambiguities in monocular views.
\par
\noindent \textbf{Learnable SoftArgmax (w/ LSA).} 
Using the learnable version of softargmax decreases MPJPE to 71.31mm, hinting that joint-specific temperature can contribute to additional accuracy.
\par
\begin{figure}[t]
\centering
\includegraphics[width=0.95\linewidth]{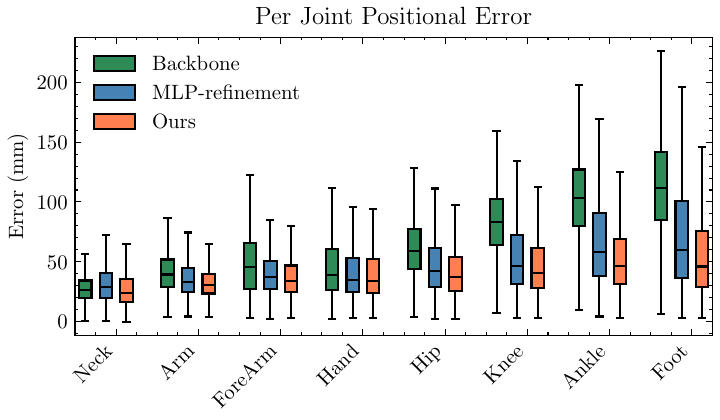}
\vspace{-3mm}
\caption{
The distribution of errors at different stages. 
In green, just after the prediction in camera coordinates, where the feet are hard to estimate. 
In blue, we can observe the improvement provided with by the rototranslation to $\Ff$ with an MLP without the time component, while in orange we see the compounded effect of both time history and floor alignment.}
\label{fig:errors}
\vspace{-5mm}
\end{figure}

\noindent \textbf{Frame Alignment.} 
One of our key contributions, i.e. leveraging the device pose by rototranslating predictions into frame $\Ff$, significantly decreases the MPJPE to 59.53mm, showing the impact the frame of reference can have. 
As shown in Fig.~\ref{fig:errors}, the most significant improvement happens in the lower limbs.
\par
\noindent \textbf{Cross Training Caching (w/ CT).} 
Introducing our Cross Training (CT) brings it down to 54.08mm suggesting that efforts in mimicking unseen data error distribution in the training set increase the ability of the model to generalize.
\par
\noindent \textbf{Stereo Temporal Fusion (STF).} 
Incorporating the STF module, which leverages previous frames information, achieves the best performance, reducing MPJPE to 47.53mm. 
This highlights the importance of temporal information, as the model can use past predictions to correct or refine current estimates.
\par 
Our ablations demonstrate that careful integration of stereo-temporal fusion coupled with the ability to work in a more meaningful frame of reference provides substantial gains, showing the value of each component in achieving state-of-the-art accuracy in egocentric pose estimation.
\input{tables/ablations}
\subsection{Generalizablity of our Floor-aligned Frame}
Our floor-aligned frame is a general design, which can be easily integrated into other egocentric pose estimation methods.
To illustrate the effectiveness and generalizability of this design, we select the monocular EgoWholeBody \cite{egowholebody} model, training it solely on images from the left camera of our dataset. 
After obtaining initial predictions, we rototranslate it to a different frame before using an MLP to refine it.
Tab.~\ref{tab:monocular} demonstrates the impact of each frame choice on MPJPE.
The baseline ($\Lf$) provides a starting point with 85.63mm. 
Aligning predictions to the middle frame ($\Mf$) yields a slight improvement, reducing MPJPE to 84.06mm. 
However, re-projecting the model's output into the floor frame ($\Ff$), i.e. our proposed design, results in a significant MPJPE reduction to 76.43mm, confirming that this alignment contributes directly to improved accuracy.
\input{tables/monocular}

%% file: tables/methods.tex
\begin{table*}[t]
\centering
\resizebox{0.95\textwidth}{!}{%
\begin{tabular}{lcccccccc} 
\toprule
Method                                  & Inference \expl{ms}{\DW} & MPJPE \expl{mm}{\DW} & PA-MPJPE \expl{mm}{\DW} & 3D-PCK \expl{\%}{\UP}  & Jitter \expl{mm}{\DW} & NPP \expl{\%}{\UP} & MPE \expl{mm}{\DW}  & FS \expl{cm/s}{\DW}         \\
\midrule                                %
Egoglass  \cite{egoglass}               & 8.97      & 105.56     & 74.11      & 61.38      &  12.60    & 52.11     & 48.16      & 12.34       \\
Unrealego \cite{unrealego}              & 6.87      & 104.81     & 68.10      & 61.22      &  11.77    & 58.03     & 48.19      & 10.71       \\
EgoPoseFormer \cite{egoposeformer}      & 14.36     & 69.18      & 41.29      & 78.98      &  9.98     & 49.45     & 47.97      & 9.29        \\
\hline
\textbf{Ours}                           & \bf{2.68} & \bf{47.53} & \bf{35.86} & \bf{92.56} & \bf{4.96} & \bf{100.0} & \bf{0.00} & \bf{3.47}   \\
\bottomrule
\end{tabular}%
}
\caption{Comparison of different egocentric models. The baselines are retrained on our dataset. Our method is the only one that leverages both the camera feeds and the device tracking. Notably, we outperform prior works by a significant margin across all metrics.}
\label{tab:methods}
\vspace{-4mm}
\end{table*}

%% file: tables/ablations.tex
\begin{table}[t]
\centering
\begin{tabular}{lcc} 
\toprule
Method                          & Frame               &  MPJPE      \\
\hline
w/o stereo                      & $\Lf$               &  87.94      \\
w/ avg                          & $\Mf$               &  84.31      \\
w/ MLP                          & $\Lf+\Rf$           &  82.19      \\
w/ MLP+avg                      & $\Mf$               &  81.69      \\
w/ MLP+avg+SH                   & $\Mf$               &  74.13      \\
w/ MLP+avg+SH+LSA               & $\Mf$               &  71.31      \\
w/ MLP+avg+SH+LSA               & $\Ff$             &  59.53      \\
w/ MLP+avg+SH+LSA+CT            & $\Ff$             &  54.08      \\
\hline
\textbf{Ours}            & $\Ff$             &  \textbf{47.53}      \\
\bottomrule
\end{tabular}%
\caption{
Ablation results. \textit{w/ avg} denotes stereo merging by averaging predictions; \textit{SH} is the Shared Head; \textit{LSA} is Learnable SoftArgmax; \textit{DN} stands for Dynamic Noise; \textit{CT} is the Cross Training caching strategy; \textit{Ours} is our Stereo Temporal Fusion module.}

\label{tab:ablation}
\vspace{-3mm}
\end{table}

%% file: tables/monocular.tex
\begin{table}[t]
\centering
\begin{tabular}{lcc} 
\toprule
Method                                 & Refinement Frame & MPJPE             \\
\midrule
EgoWholeBody \cite{egowholebody}       & $\Lf$              & 85.63             \\
EgoWholeBody \cite{egowholebody}       & $\Mf$              & 84.06             \\
EgoWholeBody \cite{egowholebody}       & $\Ff$              & 76.43             \\
\bottomrule
\end{tabular}%
\caption{Impact of the frame of reference in refinement on a monocular method trained exclusively on the left camera. Note that our proposed floor-aligned frame can signficantly improve other egocentric approaches as it is a generally applicable design.}
\label{tab:monocular}
\vspace{-4mm}
\end{table}

%% file: sec/10_conclusion.tex
\section{Discussion and Conclusion} \label{sec:conclusion}
\par
\noindent \textbf{Limitations.} 
Current VR headsets offer rich multimodal data such as environment meshes, forward-facing cameras, eye-gaze tracking, and hand pose estimation from controllers—inputs our current method does not exploit. Leveraging these modalities can further enhance the model’s ability to perceive nuanced motion and user intention.
Furthermore, while our approach robustly addresses many challenging scenarios, it can struggle with inherent motion capture issues such as significant occlusions and self-contact. Integrating physics-based modeling as a way to bridge the aforementioned multimodal data, represents a promising direction to overcome these limitations by enforcing realistic physical constraints and resolving ambiguous cases.

\par
\noindent \textbf{Conclusions.}
We presented a large-scale, real-world dataset for egocentric motion capture, surpassing existing datasets in size and motion complexity. 
This dataset includes high-quality, on-device head tracking, providing essential information for pose estimation on any device capable of tracking its 6D pose.
Our proposed method leverages known geometric transformations, such as camera and device poses, achieving precise 3D pose predictions while running efficiently at 300FPS on modern hardware. 
We demonstrated that our approach improves generalization and that our frame of reference choices significantly enhances performance over competing methods.
Experiments validate the effectiveness of our model and the architectural decisions underlying its design.
Looking ahead, we expect this motion capture model to be extended and serve as a starting point to not only estimate 3D keypoints but also physical quantities or maybe see its application in driving photorealistic avatars from egocentric signals.

\par 
\noindent \textbf{Acknowledgements.}
This work was funded by the Saarbrücken Research Center for Visual Computing and Artificial Intelligence (VIA).